\pdfoutput=1

\documentclass[11pt,a4paper]{article}
\usepackage{times,latexsym}
\usepackage{amssymb}
\usepackage{pifont}
\usepackage{placeins}
\usepackage{hyperref}
\usepackage{url}
\usepackage[T1]{fontenc}
\usepackage[acceptedWithA]{tacl2021v1}
\urldef{\umlssemtypeurl}\url{https://www.nlm.nih.gov/research/umls/knowledge\_sources/semantic\_network}
\usepackage{amsmath}
\usepackage{lineno}
\usepackage{booktabs}
\usepackage{multirow}
\usepackage{threeparttablex}
\usepackage{adjustbox}
\usepackage{bm}
\usepackage{caption}
\usepackage{subcaption}
\usepackage{tcolorbox}
\usepackage{tabularx}
\usepackage{xspace}
\usepackage{nicefrac} 
\usepackage[utf8]{inputenc}

\usepackage{microtype}

\usepackage{inconsolata}

\usepackage{graphicx}

\newcommand{\trex}{\textsc{Trex}\xspace}
\newcommand{\reveal}{\textsc{ReVeAL}\xspace}
\newcommand{\hitgnn}{\textsc{HiTGnn}\xspace}
\newcommand{\gnn}{\textsc{Gnn}\xspace}
\newcommand{\rnn}{\textsc{Rnn}\xspace}
\newcommand{\gnns}{\textsc{Gnn}s\xspace}

\title{Early Risk Prediction with  Temporally and Contextually Grounded Clinical Language Processing}

\author{
\textbf{Rochana Chaturvedi}\Thanks{This work was conducted while the author was affiliated with the University of Illinois Chicago.}\textsuperscript{1},
\textbf{Yue Zhou}\textsuperscript{2},
\textbf{Andrew Boyd}\textsuperscript{2},
\textbf{Brian T. Layden}\textsuperscript{2},
\textbf{Mudassir Rashid}\textsuperscript{3},\\
\textbf{Lu Cheng}\textsuperscript{3},
\textbf{Ali Cinar}\textsuperscript{3},
\textbf{Barbara Di Eugenio}\textsuperscript{2}
\\[0.5em]
\textsuperscript{1} Kellogg School of Management, Northwestern University, IL, USA \\
\textsuperscript{2}University of Illinois Chicago, Chicago, IL, USA \\
\textsuperscript{3}Illinois Institute of Technology, Chicago, IL, USA
\\[0.3em]
\small Correspondence: \href{mailto:rchaturvedi@anl.gov}{rochana.chaturvedi@northwestern.edu}
}

\begin{document}
\maketitle
\begin{abstract}
Clinical notes in Electronic Health Records (EHRs) capture rich temporal information on events, clinician reasoning, and lifestyle factors often missing from structured data. Leveraging them for predictive modeling can be impactful for timely identification of chronic diseases. However, they present core natural language processing (NLP) challenges: long text, irregular event distribution, complex temporal dependencies, privacy constraints, and resource limitations. We present two complementary methods for temporally and contextually grounded risk prediction from longitudinal notes. First, we introduce \hitgnn, a hierarchical temporal graph neural network that integrates intra-note temporal event structures, inter-visit dynamics, and medical knowledge to model patient trajectories with fine-grained temporal granularity. Second, we propose \reveal, a lightweight test-time framework that distills LLMs' reasoning into smaller verifier models. Applied to opportunistic screening for Type 2 Diabetes (T2D) using temporally realistic cohorts curated from private and public hospital corpora, \hitgnn achieves the highest predictive accuracy\textemdash especially for near-term risk\textemdash while preserving privacy and limiting reliance on large proprietary models. \reveal enhances sensitivity to true T2D cases and retains explanatory reasoning. Our ablations confirm the value of temporal structure and knowledge augmentation, and fairness analysis shows \hitgnn performs more equitably across subgroups.

\end{abstract}

\section{Introduction}
Modeling disease progression represents one of the most critical challenges in modern healthcare, with particularly high stakes for chronic conditions like Type 2 Diabetes (T2D), which affects half a billion people worldwide and continues rising \citep{IDF2025}. The rich information in free-text clinical notes available in EHRs provides an opportunity for NLP-based predictive modeling. These notes present two natural perspectives for understanding disease progression and suggest distinct yet complementary computational approaches. First, the \textbf{temporal progression} of symptoms, diagnoses, treatments, and interventions, which are natural for understanding the underlying disease pathology, requires structured modeling. Second, capturing the \textbf{semantic richness} of clinical notes calls for contextualized language understanding. These are particularly relevant for opportunistic screening, where patients with low socioeconomic status often miss routine care and timely screening \cite{danielson2023prevalence}. In such settings, clinicians must act on all historical information available at a given visit to estimate the disease risk. Formally, given patient documents ${d_1, d_2, \dots, d_n}$ ordered in a non-decreasing sequence of patient visits ${1, 2, \dots, n}$, we want to estimate the risk at next visit ($n+1$). 

To address these complementary modeling needs, this work proposes two novel methods that integrate domain-specific modeling with targeted representation learning techniques over clinical notes, optimized for low-resource settings. To address structured temporal modeling, we propose \textbf{Hi}erarchical \textbf{T}emporal \textbf{G}raph \textbf{N}eural \textbf{N}etwork (\hitgnn), a dynamic model that captures patient state evolution with fine-grained temporal granularity and knowledge-enhanced structure. This fine-grained knowledge is crucial for understanding disease pathways. For example, elevated glucose \textbf{After} corticosteroid use likely indicates drug-induced hyperglycemia, not T2D, while the same glucose levels \textbf{Before} any steroid therapy may signal early T2D onset. This motivates our \hitgnn framework, which leverages temporal graphs extracted from individual clinical notes using clinical temporal relation extraction (\trex) methods. \trex involves identifying entities such as clinical events (problems, treatments), time expressions or timex (e.g., dates), and their temporal relations (Before, After, and Overlap). \hitgnn augments these graphs with semantic information from a clinical knowledge base. These multi-layered event-temporal graphs are modeled with dynamic graph neural networks (\gnns) to capture patient timelines within and across visits. While \gnns are effective for relational reasoning, their performance depends on rich contextual embeddings; thus, we incorporate clinically pretrained language model (CPLM) embeddings and knowledge-graph embeddings to enhance input representations with rich semantics. This addresses the long-recognized need to incorporate causal and temporal patterns into diagnostic reasoning \citep{patil1981causal}. 

Complementing this structured approach, current Large Language Models (LLMs) offer strong localized and context-aware reasoning capabilities to model the underlying text semantics. However, they often struggle with long-range context integration and structured reasoning\textemdash limitations particularly pronounced in the clinical domain, where patient histories span multiple long notes. Additionally, real-world healthcare deployments face privacy and resource constraints, limiting the use of proprietary or large-scale models. Our second method, \textbf{Re}asoning with \textbf{Ve}rifier-\textbf{A}ided \textbf{L}abeling (\reveal), is a scalable inference-time architecture preserving LLM-style reasoning in resource-constrained settings. It distills reasoning traces from a large LLM into a smaller one, scaling performance with interpretability without the full computational cost of training larger models. 

Together, these complementary modeling paradigms\textemdash graph-based temporal reasoning and interpretable LLM inference\textemdash provide a holistic risk prediction framework grounded in structure and semantics. We evaluate them in the context of opportunistic screening of T2D as a representative use case, using a real private hospital (PH) corpus and a corpus curated from MIMIC-IV \citep{johnson2023mimic}. Additionally, due to the prevalent demographic biases \citep{meng2022interpretability,zhou2025unveiling, hall2015implicit}, we conduct a fairness analysis to better understand model behavior across demographic groups.

 \textbf{Contributions. } (1) We present two complementary representation learning frameworks from longitudinal clinical notes, tailored for low-resource settings:
 (i) \textbf{\hitgnn}: the first application of clinical temporal relation extraction (\trex) for risk prediction that integrates intra-document temporal relations, inter-visit dynamics, and medical knowledge, enabling reasoning across both local event structures and longitudinal patient trajectories.
 (ii) \textbf{\reveal}: an inference-time scaling framework where a smaller LLM validates predictions from a larger frozen LLM, inheriting interpretability and improving accuracy without full retraining.\\
(2) We demonstrate the translational value of temporally enriched representations in a real clinical application\textemdash opportunistic screening for T2D, especially in immediate-risk horizons, where intervention is most impactful.\\
(3) We curate rigorous datasets from private (PH) and public (MIMIC-IV) sources, excluding post-diagnosis inputs to prevent label leakage, and ensuring balanced cohorts for fair evaluation.\\
(4) Extensive ablations disentangle the impacts of temporal relations, KG augmentation, representation enrichment, and pipeline components, revealing corpus-dependent trade-offs.\\
(5) Our fairness analysis highlights how LLM-based systems exhibit instability and inconsistency in subgroup behavior, even with demographic conditioning, and that \hitgnn generally is fairer even without explicitly encoding demographic attributes.\\
These contributions establish a clinically grounded framework integrating concept abstraction, temporal structure, and semantic enrichment to support low-cost and privacy-conscious AI systems for real-world clinical decision support. 

\section{Related Literature}
\paragraph{Computational Approaches for Disease Progression Prediction}
Despite significant advances in predictive modeling using electronic health records (EHR), many studies continue to rely predominantly on structured EHR data \citep{lipton2015learning,singhal2023detect, jianggraphcare}, which is often noisy \citep{hersh2013caveats} and incomplete \citep{capurro2014availability}. Among efforts that leverage free-text, recent works either use only the last clinical note \citep{xu2023vecocare, nguyen2024carer}, relying on structured data to capture temporal trends; or use a coarser form of temporal modeling where each note is embedded using a language model \cite{huang2020clinical}, a bag-of-concepts \citep{chaturvedi2023sequential}, or topics \citep{ghassemi2015multivariate}, and a BiLSTM models the representations from multiple notes across patient visits. While these studies explore various fusion strategies, a key gap remains: no prior work models temporal relations both within a single clinical note and across multiple notes/visits to forecast long-term disease risk. Our work directly addresses this gap by:
(1) leveraging the sequence of clinical notes for each patient, and
(2) introducing multi-layered modeling to leverage temporal information extracted from each note and across the notes from multiple visits to track evolving health trajectories.

\paragraph{LLM in Healthcare}
The integration of Large Language Models (LLMs) into healthcare is a rapidly evolving field \cite{zhou2023survey,he2024survey}. Key developments include impressive performance on medical question answering (QA) benchmarks \citep{singhal2023towards}. However, outside of controlled QA settings, LLMs still struggle on tasks requiring complex reasoning for disease diagnosis in real clinical settings \citep{hager2024evaluation,wang2024assessing}, and often exhibit performance disparities across demographic groups \citep{zhou2025unveiling}. Recent research has demonstrated that scaling test-time compute through verifier-guided search can significantly improve LLMs' reasoning capabilities and often outperforms simply scaling model size \citep{snell2024scalingllmtesttimecompute}. However, scaling test-time compute of LLMs in low-resource real-world healthcare tasks is still underexplored. Our work contributes to this literature by exploring verifier-guided reasoning approaches tailored to low-resource clinical prediction tasks.

\section{Data Curation}
\label{sec:data}
\begin{figure*}[htbp]
    \centering    \includegraphics[width=\linewidth,  trim=0 10 0 0, clip]{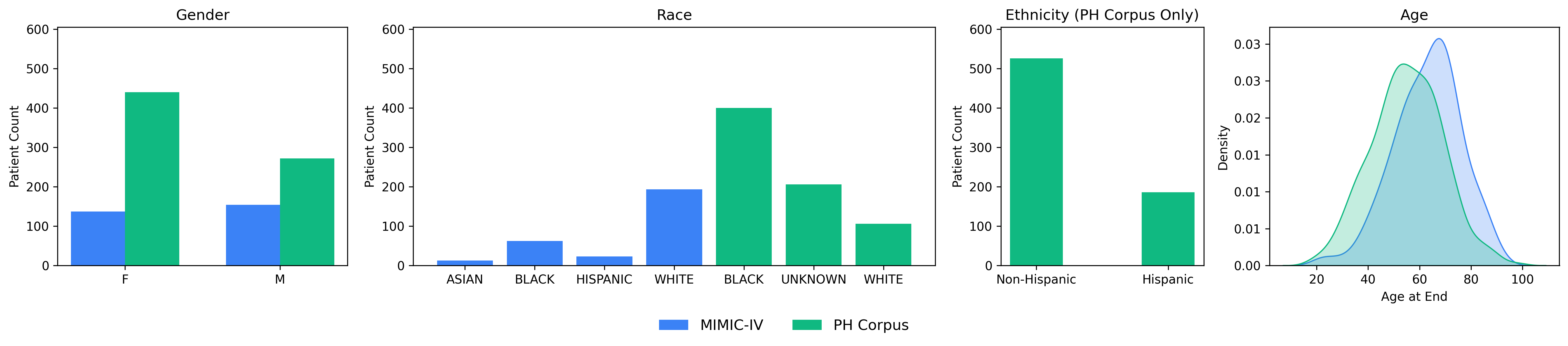}
    \caption{Demographic distribution in PH and MIMIC-IV test sets.}
    \label{fig:dem}
      \vspace{-3mm}
\end{figure*}
 Our first dataset, the PH corpus, is curated from private data comprising adult patients (age $\geq$18 years) collected from a U.S.-based hospital between January 2010 and July 2021. We also curate a second corpus from MIMIC-IV (Medical Information Mart for Intensive Care, version 4) to study the generalizability of our methods. MIMIC-IV is a large, publicly available database comprising de-identified clinical notes from ICU encounters at Beth Israel Deaconess Medical Center in Boston from 2008\textendash2019. We exclude diagnosis of other types of diabetes (e.g., Type-1 diabetes, gestational diabetes) using ICD codes (standardized diagnostic codes) to avoid conflating related yet distinct diagnostic trajectories. Therefore, our results focus on distinguishing chronic adult T2D\textemdash which accounts for 90--95\% of diagnosed adult diabetes cases \citep{CDC}\textemdash from non-diabetes (NoD). 
 
For patients in the NoD group, we include all available clinical notes except the last recorded encounter in the dataset, which is indicative of no diabetes. For the T2D cohort, an additional refinement step ensures that post-diagnosis data is excluded to prevent label leakage. While we use the date of the first recorded ICD code as a proxy for Type 2 Diabetes (T2D) onset, this signal is known to be noisy \citep{hersh2013caveats} and often delayed relative to when the diagnosis is first documented in clinical notes. To mitigate this, we additionally process the T2D cohort's notes using a large language model (LLM) to identify the earliest explicit mention of a T2D diagnosis in unstructured text. This kind of additional filtering is crucial for ensuring realistic model evaluation. For example, \citet{zhang2024rethinking} show that a widely used sepsis prediction model frequently issues risk alerts only after clinicians have documented suspicion, undermining its practical utility. A substantial fraction of patients show a mismatch between the recorded diagnosis date and an explicit textual mention of T2D across PH ($585/2{,}302\approx25\%$) and MIMIC-IV ($2{,}298/5{,}207\approx44\% $). Manual review of 150 notes from 30 randomly sampled patients in the T2D cohort of the final datasets found 1 violation in the PH corpus and no violations in MIMIC-IV.\footnote{The LLM frequently handles borderline cases by correctly inferring the absence of a T2D diagnosis from contexts such as negated mentions (e.g., ``denies DM''), risk states (``pre-DM'', ``borderline DM''), action plans (``DM screening'', ``A1c ordered''), family history of diabetes, and alternate medication use (e.g., ``metformin'' prescribed for ``morbid obesity'').} Review of excluded notes shows perfect accuracy in identifying prior or current T2D diagnoses.

Finally, we construct a demographically matched test set for fair evaluation. We estimate T2D propensity scores \citep{rosenbaum1985constructing} from demographic variables, including age, gender, and race (and ethnicity in the case of the PH corpus; this is combined with Race in MIMIC-IV). This is followed by 1:1 greedy nearest-neighbor matching without replacement to select NoD controls.\footnote{The data filtering details are provided in Appendix \ref{app:data}.} The final PH corpus comprises 3332 patients (712 in the test set), and the final MIMIC-IV subset contains 5802 patients (291 in the test set). 
 
 Both datasets pose unique strengths and limitations: PH contains richer longitudinal histories with diverse note types, ideal for chronic disease modeling, but it is not public. In contrast, while MIMIC-IV is one of the most widely used, large public datasets, it only focuses on ICU notes, omitting much of the hospitalization timeline, and has narrower temporal windows (84.5\% patient records comprise single visits \citep{cui2025timer}). Further, MIMIC-IV notes are de-identified, lacking an essential temporal anchor\textemdash DATE\textemdash considered protected health information (PHI).\footnote{Health Insurance Portability and Accountability Act (HIPAA), a U.S. law, considers dates as PHI. In MIMIC-IV notes, all PHIs are identified by a union of a rule-based and a neural approach, and replaced with three underscores `\_\_\_' \citep{johnson2023mimic}.} Figure \ref{fig:dem} shows demographic differences: \textit{females} are the majority in PH but a minority in MIMIC-IV; \textit{Black} is the majority in PH, \textit{White} in MIMIC-IV, with Asian as the minority. PH corpus records Hispanics under a separate ethnicity attribute. Therefore, evaluating methods on both datasets is important. However, with MIMIC-IV's limited temporal continuity, performance on this dataset likely represents a lower bound compared to richer data settings.

\section{\hitgnn: Disease Prediction using Hierarchical Temporal Graphs}
\label{sec:proposedfinegrained}

\begin{figure}
    \centering\includegraphics[width=\linewidth]{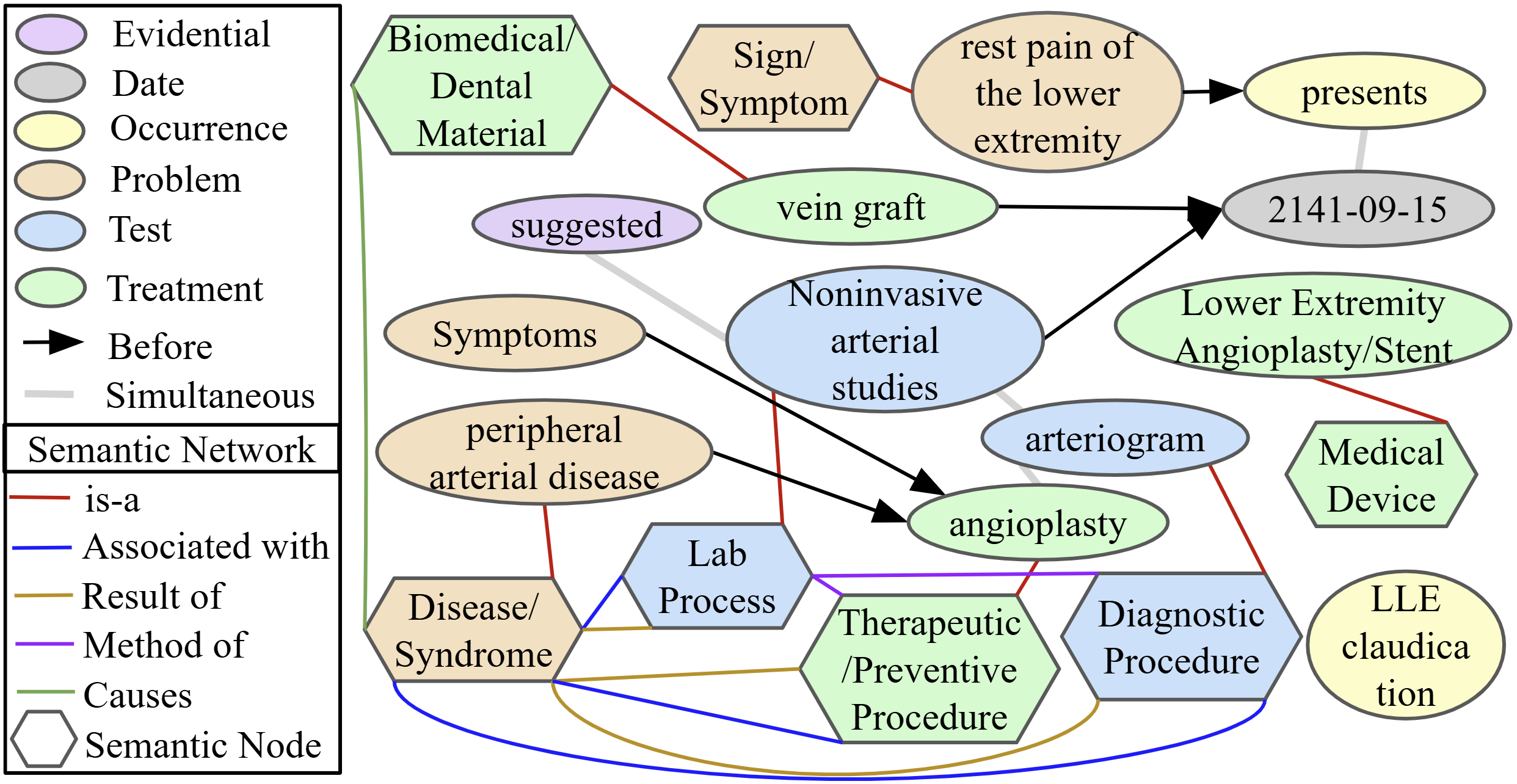}
    \caption{An example of an extracted graph from one clinical note of MIMIC-IV. Extracted entities (clinical events and time expressions) are circular nodes, color-coded by type, and have temporal edges (Before, Simultaneous). This is further augmented with the UMLS KG, which includes semantic nodes connected to the corresponding entity via the \textit{is-a} relation and to each other via semantic relations (e.g., \textit{causes}, \textit{method of}).}
    \label{fig:tkg_example}
    \vspace{-3mm}
\end{figure}
\hitgnn is a \textbf{Hi}erarchical \textbf{T}emporal \textbf{G}raph \textbf{N}eural \textbf{N}etwork for modeling fine-grained temporal information and external knowledge to predict T2D risk. We extract and refine temporal graphs from clinical notes (sections \ref{sec:extract_tkg}\textendash\ref{sec:reducedgraph}, example in Figure \ref{fig:tkg_example}), and initialize graph nodes with complementary signals (Section \ref{sec:node_init}) before modeling them across visit sequence (Section \ref{sec:hi-temp}).
\subsection{Extracting and Aligning Event-Temporal Graphs from Notes}
\label{sec:extract_tkg}
We use state-of-the-art models from \citet{chaturvedi-etal-2025-temporal} to extract temporal graphs from each note. Here, nodes are clinical entities (\textit{Problem, Test, Treatment, Clinical Department, Clinical Occurrences, Evidential}), time expressions (\textit{Date, Time, Duration, Frequency})\textemdash circular nodes in Figure \ref{fig:tkg_example}\textemdash and edges, their temporal relations (\textit{Before}, \textit{After}, \textit{Overlap/Simultaneous}). We \textit{align} extracted entities by linking and clustering multiple mentions into a single node (Section \ref{sec:linking}). We optionally denoise temporal edges by enforcing consistency constraints using TimeGraph and normalizing directed edges (Before/After) to Before (Section \ref{sec:reducedgraph}).

\subsection{Entity Linking/Normalization} \label{sec:linking}

\paragraph{Clinical Entities} Next, we perform clinical entity alignment by clustering synonymous textual mentions into a single node via linking to an external knowledge graph. If any entity is of type \textit{Problem/Treatment/Test/Clinical Department}, we use Metamap  \citep{aronson2010overview} to map it to a unique concept in the Unified Medical Language System (UMLS) Metathesaurus \citep{bodenreider2004unified}.\footnote{We do not link other entity types, as it leads to noise (Appendix \ref{app:ph_annotation}).} UMLS comprises a vast repository mapping synonymous medical terms to unique concept identifiers (CUI). Additionally, UMLS defines a Semantic Network that provides a higher-level abstraction over individual concepts by organizing them into coarser semantic types (e.g., Disease or Syndrome, Diagnostic Procedure, Medical Device). For each extracted concept, we identify its corresponding semantic type using MetaMap and add it as an additional node in the graph (hexagonal nodes in Figure \ref{fig:tkg_example}). A set of permissible semantic relations between semantic types is also pre-specified, which we incorporate as edges in the graph (e.g., Diagnostic Procedure \textit{associated with} Disease or Syndrome, shown as a blue edge in Figure \ref{fig:tkg_example}). Each concept entity node is connected to its associated semantic type node via \textit{is-a} relations (red edges in Figure \ref{fig:tkg_example}; e.g., stent \textit{is-a} Medical Device), explicitly modeling the type hierarchy.\footnote{The complete set of semantic types and relations is available at: \umlssemtypeurl.} This yields a heterogeneous graph combining extracted temporal graphs with medically grounded abstractions.
\paragraph{Dates} We normalize and align multiple textual mentions of the same date to a standard representation using Microsoft Recognizers-Text \citep{soft-recognizers-text}. All mentions of a date entity are clustered as a single node in the temporal graph.

\subsection{Consistent Temporal Graph }
\label{sec:reducedgraph}
We additionally evaluate temporal edge denoising via consistency enforcement using the TimeGraph algorithm \citep{miller1990time}. This algorithm iteratively adds the most confident edges based on prediction probabilities to the final graph, and prunes those that introduce cycles. The resulting graph contains \textit{Before}, \textit{After}, and \textit{Simultaneous} edges (black and grey edges in Figure \ref{fig:tkg_example}). Following prior observations that temporal relations form inverse, redundant pairs \citep{mani2006machine}, and inspired by work emphasizing canonical temporal ordering via \textit{Before} relations \citep{chambers2008unsupervised}, we flip the direction of \textit{After} edges and relabel them as \textit{Before}. This reduces label sparsity and simplifies temporal supervision without imposing additional modeling constraints. Table \ref{tab:graph-stats} presents descriptive statistics of the final graphs.\footnote{The mean number of tokens to get estimated document length is counted based on Llama3.1-8B tokenizer, as we also include comparisons with this model.}

\subsection{Node Representation}\label{sec:node_init}
To represent a node, we compute its corresponding textual span's embedding by concatenating the in-context BioMedBERT \citep{gu2021domain} embeddings of the first and last tokens with the span-width embeddings. This model is pretrained on biomedical corpora and demonstrated strong performance on domain-specific tasks.\footnote{We use the fine-tuned model version and width embeddings from \citet{chaturvedi-etal-2025-temporal}.} We represent multiple linked mentions of an entity as a single node in each temporal graph, and initialize it with the mean of the embeddings of all linked mentions. To represent semantic type nodes, we average BioMedBERT embeddings over all tokens in their label (preferred name). We also use knowledge graph embeddings (KG-embeddings) for Concept Unique Identifiers (CUIs) derived from the UMLS knowledge graph for each linked event, as introduced by \citet{maldonado2019adversarial}. The final node representations are concatenated KG embeddings and text embeddings (BioMedBERT) over all mentions.

\begin{table}[htbp]
    \centering
    \resizebox{.9\linewidth}{!}{
    \begin{tabular}{lcc}
    \toprule
\textbf{Parameter}&\textbf{PH Corpus} &\textbf{MIMIC-IV}\\
\midrule
\textbf{\#Patients}&3332&5802\\
\textbf{Avg. \#tokens/doc}&1209.2 (637.6) &454.5 (208.3)\\
\textbf{Avg. \#nodes/doc}&114.7 (46.1)& 74.8 (27.9)\\
\textbf{Avg. \#Edges/doc}   &   261.1 (133.4)& 158.2 (76.8)\\
\textbf{\#doc or visits}& 3.2(1.7)&2.3 (1.4)\\
\bottomrule
    \end{tabular}}
    \caption{Mean tokens, nodes, and edges in the extracted temporal graph per document (doc), per patient, and number of documents per patient, with standard deviations in parentheses.}
     \label{tab:graph-stats}
         \vspace{-3mm}
\end{table}
\subsection{Hierarchical Temporal Modeling}
\label{sec:hi-temp}
\begin{figure*}
 \centering    \includegraphics[width=.95\linewidth,  trim=13 50 0 90, clip]{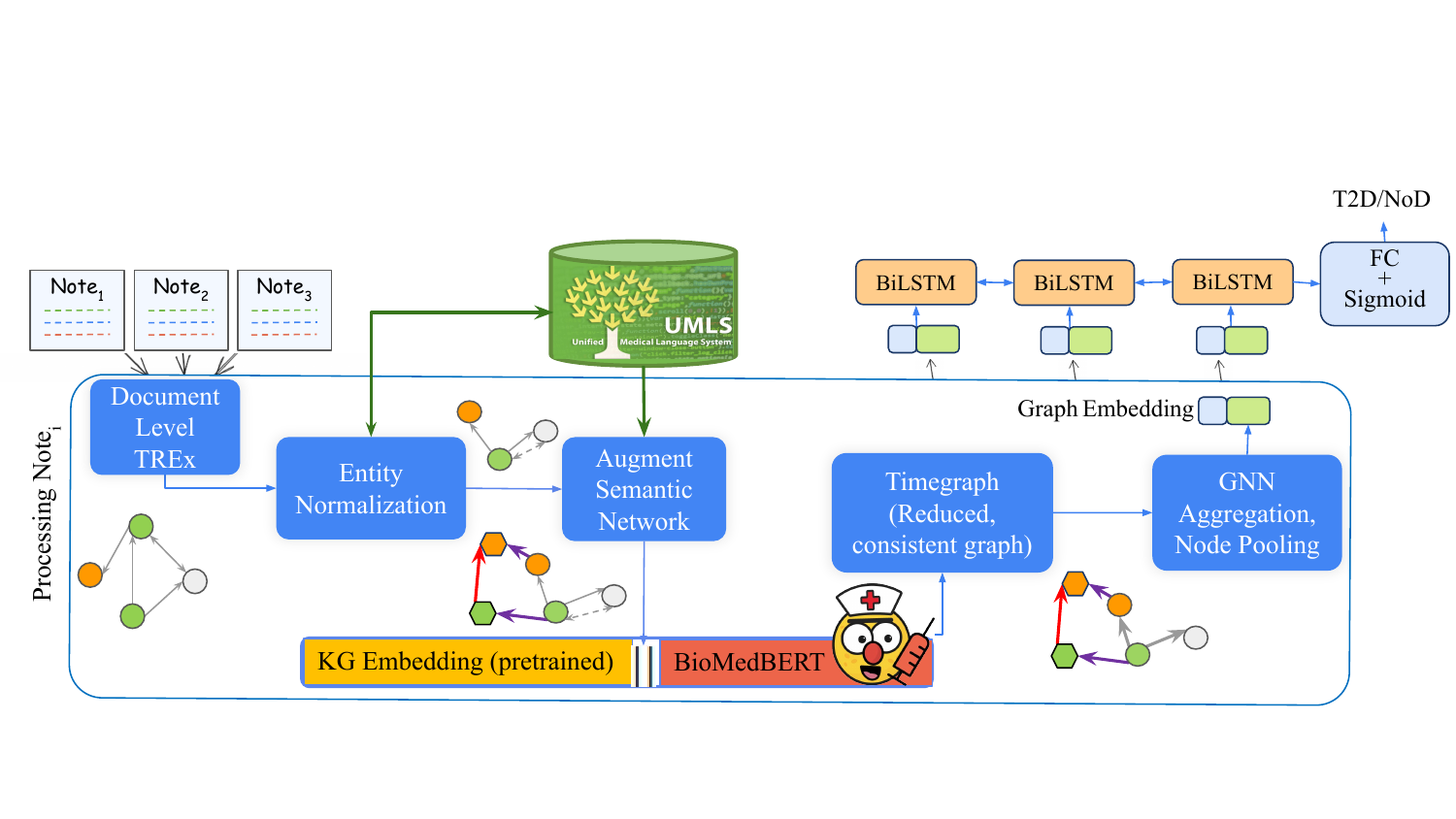}
 \vspace{-2mm}
    \caption{\hitgnn architecture for longitudinal modeling of clinical trajectories, combining document-level temporal graphs with cross-visit aggregation and UMLS knowledge for T2D risk prediction.}
    \label{fig:temporal_gnn_encoder}
\end{figure*}
We model irregular time-series of temporal graphs across patient visits using \hitgnn for Type 2 Diabetes prediction. Each visit is a graph $\mathcal{G}_t = (\mathcal{V}_t, \mathcal{E}_t)$ with temporal (Before/After/Overlap) and semantic (UMLS) edges. \hitgnn captures intra-graph and inter-graph dependencies via a \gnn and a recurrent neural network (\rnn).

\textbf{Intra-visit Modeling}. Each graph is encoded with a multi-layer GraphSAGE encoder (with mean aggregation, residual connections, and layer normalization). Given initial features $\mathbf{h}_v^{(0)}$ of node $v$, $\mathcal{N}(v)$ the neighbors of $v$, and learnable parameters $\mathbf{W}_1^{(k)}, \mathbf{W}_2^{(k)}$, node updates at layer $k$ are:
\begin{align*}
\mathbf{h}_v^{(k)} &= \text{LayerNorm}\Big(\mathbf{W}_1^{(k)} \mathbf{h}_v^{(k-1)} \notag\\
&\quad + \mathbf{W}_2^{(k)} \frac{1}{|\mathcal{N}(v)|}\sum_{u \in \mathcal{N}(v)} \mathbf{h}_u^{(k-1)}\Big) + \mathbf{h}_v^{(k-1)}
\end{align*}
\noindent Where \( \mathbf{h}_v^{(k)} \) is the representation of node $v$ at layer  $k$.
Updated node embeddings are mean-pooled to obtain the graph representation for a visit $\mathbf{g}_t$.

 \textbf{Inter-visit Modeling}: temporal dependencies across visits are modeled as:
$$
\mathbf{z}_t = \text{BiLSTM}_\phi(\mathbf{g}_1, \dots, \mathbf{g}_t), \qquad
\mathbf{z}_T = [\overrightarrow{\mathbf{h}}_T \, ; \, \overleftarrow{\mathbf{h}}_1]
$$
Where $\mathbf{g}_1, \dots, \mathbf{g}_t$ is the sequence of graph encodings over visits. The final forward and backward hidden states from bidirectional LSTM ($BiLSTM$) are concatenated 
to represent a patient's trajectory. Finally, a fully connected network with sigmoid ($\sigma(\cdot)$) gives binary outcome ($\hat{y}$)\textemdash T2D or NoD (No Diabetes): $$\hat{y} = \sigma(\mathbf{W} \cdot \mathbf{z}_T + \mathbf{b})$$

Figure \ref{fig:temporal_gnn_encoder} shows the model architecture, integrating UMLS knowledge graph, document-level temporal information, and cross-visit progression.

\section{LLMs for Diabetes Risk Prediction}
In this section, we use Large Language Models (LLMs) as reasoning engines to predict a patient's risk of developing T2D from the clinical notes.

\subsection{Reasoning with Verifier-Aided Labeling (\reveal)}
Healthcare applications face significant resource constraints, which limit the deployment of LLMs; yet, maintaining reasoning capabilities is crucial for effective clinical decision-making. Recent advances in test-time compute scaling offer a promising solution: using smaller models enhanced by verification mechanisms, rather than relying solely on larger, computationally expensive ones. To this end, the \reveal framework combines a reasoner LLM with a smaller verifier LLM, in three steps: (1) a reasoner LLM generates $N$ reasoning paths with predictions; (2) a fine-tuned verifier evaluates the credibility of these paths and assigns a score from 0\textendash 1, and (3) predictions across different reasoning paths are aggregated using verifier scores.

Concretely, at \textit{training time}, the reasoner model generates five stochastic outputs per example, each including a prediction (`true'/`false') and an explanation. A small LLM verifier, fine-tuned with LoRA \citep{hu2022lora}, classifies each output as `correct' or `incorrect' using the prompt in Figure \ref{fig:reveal_prompts}. The normalized probabilities of the output tokens serve as the verifier's confidence score. At \textit{inference time}, the reasoner generates $N$ ($N=10$) diverse predictions. Each prediction contains a risk outcome (`true'/`false') and a corresponding explanation. The verifier scores them, and the outcome is selected by majority vote among the top-$k$ highest-confidence predictions.

\section{Experiments}
This section presents a comprehensive empirical evaluation of \hitgnn and \reveal.\footnote{See Appendix \ref{app:implementation} for prompts and experimental details.}
\subsection{Code and Data} The replication code is available at \url{https://github.com/RochanaChaturvedi/temporal-nlp-clinical-risk}. We also release non-identifiable metadata to reconstruct the MIMIC-IV T2D cohort, including patient identifiers, note timestamps, and demographic attributes. The full MIMIC-IV-T2D dataset will be released via PhysioNet \citep{goldberger2000physiobank}.

\subsection{Baseline Methods}
\noindent \textbf{Zero-shot Prompting}: The model directly predicts T2D risk from clinical notes using deterministic decoding (temperature=0.0). We extract normalized token probabilities for `true'/`false' predictions to compute AUC scores (see prompt in Figure \ref{fig:Llama_dp_reasoner_prompt}). We use Llama3.1-8B and Llama3.2-1B, and also compare the performance on MIMIC-IV with prevailing large-scale models, DeepSeek-V3  (671B parameter Mixture-of-Experts model) and GPT-4o.\footnote{Not evaluated on PH corpus due to privacy constraints.} 

\noindent \textbf{Self-Consistency (Llama3.1-8B-SC)}: We generate $N$ ($N=10$) diverse predictions per patient using stochastic sampling (temperature=1.0) and take the majority vote for the final classification.

\noindent \textbf{Supervised Fine-tuning (SFT)}: We fine-tune a 1B LLM using LoRA given the low-resource setting. The resulting model (Llama3.2-1B-ft) is trained to generate only `true' or `false' labels, without explanations, due to the lack of annotated reasoning data (see prompt in Figure \ref{fig:Llama_dp_finetune_prompt}). 

\noindent \textbf{CLSTM} We also use concept-based CNN-LSTM baseline \citep{chaturvedi2023sequential}. Here, we model extracted entities (nodes of temporal graphs) with a CNN and max pooling, then process each document representation with an LSTM. 

Both CLSTM and \hitgnn models utilize \textsc{SpanTREx}\textemdash a leading open-source end-to-end temporal relation extraction model from \citet{chaturvedi-etal-2025-temporal}. We also compare with their best model \textsc{GraphTREx} in section \ref{sec:trexablation}.

\subsection{Evaluation Metrics}
Both datasets are constructed with balanced case-control sampling to ensure sufficient representation of T2D patients. Consequently, data distribution does not reflect real-world prevalence ($\approx 11.6\%$ in the U.S.); therefore, absolute precision values may be optimistic. All models are evaluated under identical conditions; thus, relative comparisons remain valid. We emphasize ROC-AUC  and T2D recall as prevalence-independent measures of discrimination performance.
For completeness, we also report precision and recall for both groups, and the macro-average $F_1$.\footnote{ROC-AUC scores are computed from prediction probabilities. For LLMs, this is the probability of generating `true' at the predicted token position. For LLM-SC, it is the `true' token probability, averaged across all runs.} We also use the following metrics for fairness evaluation:
\paragraph{Demographic Parity Difference (DPD)} difference in the probability of positive prediction rates between a target group ($Z=z$) and the other groups ($Z \ne z$). It is defined as:
$$\text{DPD} = P(\hat{Y} = 1 \mid Z = z) - P(\hat{Y} = 1 \mid Z \ne z)$$
\noindent Where $\hat{Y}$ is the prediction (T2D $=1$, NoD $=0$), and $Z$ is a sensitive attribute (e.g., gender, race).

\paragraph{Equal Opportunity Difference (EOD)} difference in true positive rates ($TPR$) between the target group and others: $TPR_{Z = z} - TPR_{Z \ne z}$

\noindent Where $TPR_{Z = z} = P(\hat{Y} = 1 \mid Y = 1, Z = z)$.

\section{Results and Discussion}

\begin{table*}[h]      
    \centering
    \adjustbox{max width=.9\textwidth}{
    \begin{tabular}{l|cc|cc|cc|cc|cc|cc}
    \toprule\multirow{3}{*}{\textbf{Model}}&\multicolumn{6}{c|}{\textbf{PH Corpus}}&\multicolumn{6}{c}{\textbf{MIMIC-IV}}\\
    \cline{2-13}
    &\multirow{2}{*}{\textbf{AUC}}&\multirow{2}{*}{$\mathbf{F_1}$}&\multicolumn{2}{c|}{\textbf{T2D}}&\multicolumn{2}{c|}{\textbf{NoD}}&\multirow{2}{*}{\textbf{AUC}}&\multirow{2}{*}{$\mathbf{F_1}$}&\multicolumn{2}{c|}{\textbf{T2D}}&\multicolumn{2}{c}{\textbf{NoD}}\\
        \cline{4-7}\cline{10-13}
         &  &&P & R &P & R& &&P & R&P & R \\
        \midrule 
        \textbf{Llama3.2-1B (0-shot)}&  46.80& 35.46& 39.29 &3.09 & 49.56&    \textbf{95.22}&40.00& 33.10&0  &  0&49.48  &  1.00\\
        \textbf{Llama3.1-8B (0-shot)} & 57.73
        &56.98 &57.49& 53.93 &56.61  &60.11  
        &53.71
         &51.03&61.54  &27.21&52.65& 82.64\\
        \textbf{DeepSeek-V3 (0-shot)}&$^{\dagger}$\textsc{n/a}&\textsc{n/a}&\textsc{n/a}&\textsc{n/a}&\textsc{n/a}&\textsc{n/a}&55.02&47.85& \underline{69.23}& 18.37&52.38& 91.67\\
\textbf{GPT-4o  (0-shot)}&\textsc{n/a}&\textsc{n/a}&\textsc{n/a}&\textsc{n/a}&\textsc{n/a}&\textsc{n/a}&57.75 &51.81 &\textbf{75.56}  &23.13 & 54.07 & \underline{92.36}\\
        \textbf{Llama3.1-8B-SC}&55.62
        &55.49 &56.29   & 50.28 &55.08 &60.96 
        &52.12
        &43.43&66.67&12.24&51.14& \textbf{93.75}  \\
         
         \textbf{Llama3.2-1B-ft}&65.06 
         &61.62 & 59.53 &  \textbf{78.93} &  \underline{68.75}& 46.35
         &\underline{64.41} &59.17&63.89 & 46.94 & 57.38 & 72.92 \\
         
         \textbf{\reveal} &  57.98&65.24&\underline{66.77}&60.96&64.08 &69.66
        & 53.13&51.60&55.21&36.05&51.79 &70.14\\
         \midrule
         \textbf{CLSTM}&\underline{68.24}&\underline{65.62}&\underline{63.19}&\underline{76.69}&\textbf{70.36}&55.34
        &63.08&\underline{59.62}&63.96 & \underline{48.30}&\underline{57.78} &72.22\\  
        
        \textbf{\hitgnn} &\textbf{72.24}& \textbf{67.28} &\textbf{71.48} & 58.43 & 64.85 & \underline{76.69} &  \textbf{67.49}&\textbf{61.87}&65.81&\textbf{52.38}&\textbf{59.77}&72.22\\
        \midrule
         \textbf{\#patients}&\multicolumn{2}{c|}{712}&\multicolumn{2}{c|}{356}&\multicolumn{2}{c|}{356}&\multicolumn{2}{c|}{291}&\multicolumn{2}{c|}{147}&\multicolumn{2}{c}{144}\\
        \bottomrule
    \end{tabular}}
    \caption{Performance of \hitgnn and \reveal against different baselines. $^\dagger$Evaluation of external models on private data is not possible due to privacy restrictions on identifiable data.}    
    \label{tab:dp_results}
    \vspace{-3mm}
\end{table*}
Table \ref{tab:dp_results} presents the main experimental results.\footnote{We also perform a pairwise comparison of the top four models using bootstrap sampling to estimate whether the differences are statistically significant, based on non-parametric bootstrap resampling test (see Appendix Table \ref{app:significance}).} The first panel comprises LLM-based models. Among zero-shot LLM results, Llama3.2-1B fails to identify patients at risk of diabetes, classifying almost all patients into the `no-risk' (NoD) cohort. The performance improves with the higher capacity 8B variant. While self-consistency (SC) has shown promising performance in other tasks and domains, Llama3.1-8B-SC performs poorly on both datasets. The Llama3.2-1B-ft (fine-tuned) demonstrates strong performance, achieving an AUC of 65.06\% on PH and 64.41\%\textemdash the second highest among all models\textemdash on MIMIC-IV. While this fine-tuned LLM excels as a classifier, it cannot provide reasoning or explanations for its outputs.\footnote{Obtaining gold-standard reasoning data at scale for fine-tuning is costly and labor-intensive, especially from already overburdened clinical experts, limiting the feasibility of fine-tuning with reasoning.} Our \reveal model does not significantly outperform the Llama3.2-1B-ft. However, it preserves the reasoning output and improves upon the zero-shot 8B variant in terms of T2D group recall, highlighting its sensitivity in capturing true positives. For larger models, including DeepSeek-V3 with impressive clinical decision-making capabilities \citep{sandmann2025benchmark}, the gains are less pronounced: its AUC is similar to the much smaller Llama3.1-8B, but with lower T2D recall. GPT-4o attains a higher AUC but exhibits worse T2D recall than both Llama3.1-8B and \reveal, and an overall weaker performance than Llama3.2-1B-ft.

The second panel summarizes results with structured approaches. The CLSTM model provides a strong baseline, showing higher (on PH) or comparable discrimination (on MIMIC-IV) to LLMs. \hitgnn performs even better, attaining significantly higher AUC on the PH corpus, compared to other top-performing models (\reveal, Llama3.2-1B-ft, and CLSTM). On MIMIC-IV, it also attains significantly higher AUC over \reveal and CLSTM, and significantly higher T2D recall than \reveal and Llama3.2-1B-ft. In contrast, no other model shows significant improvements over \hitgnn across any metric on either corpus. \hitgnn also outperforms both the larger models (DeepSeek-V3 and GPT-4o) across key metrics. We next present the results from additional analyses and ablations.

\subsection{Prediction Horizons}

\begin{figure*}[!htbp]
    \centering    \includegraphics[width=1\linewidth,  trim=3 12 0 0, clip]{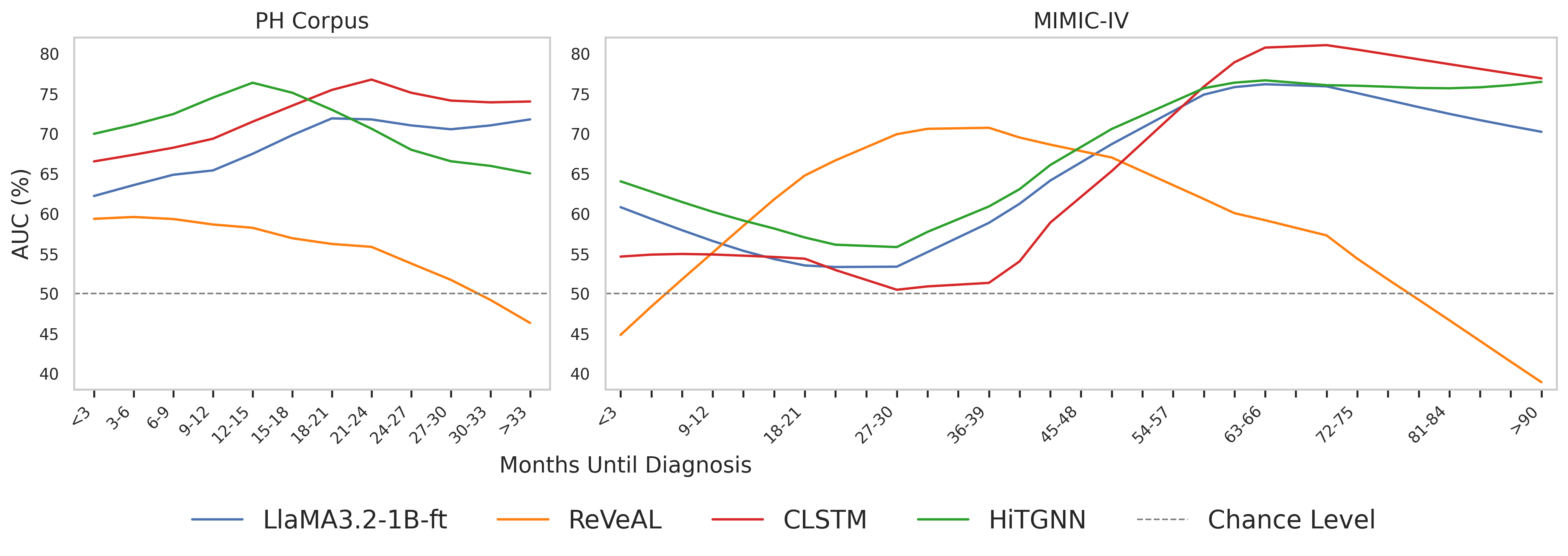}    
    \caption{AUC as a function of the prediction horizon, evaluated over consecutive 3-month windows.}
    \label{fig:dtd-ablation}\vspace{-3mm}
\end{figure*}
We compare performance across prediction horizons by dividing the time from last pre-diagnosis visit to diagnosis into 3-month windows, combining each T2D subgroup with all NoD controls to compute AUC (Figure \ref{fig:dtd-ablation}).\footnote{Appendix \ref{app:ablation_recall} Figure \ref{fig:dtd-ablation_recall} shows trends for T2D group recall.} The final window aggregates values above the $95^{th}$ percentile to avoid sparsity. Results for the four best-performing models (\hitgnn, CLSTM, \reveal, and Llama3.2-1B-ft) are presented. Across corpora, \hitgnn outperforms other models in near-term prediction ($\le1.5$ years), supporting use cases centered on more urgent risk stratification and timely preventive interventions, and reflects practical EHR constraints, where longitudinal completeness is rare. While CLSTM is not as effective in the short term, at longer horizons, it consistently achieves the highest AUC. Llama3.2-1B-ft also exhibits improvements over the longer term.  \hitgnn exhibits smoother performance degradation in MIMIC-IV and begins to match CLSTM in the long term. While \reveal performance on PH shows consistent degradation over time, on MIMIC, it initially is low but achieves the best AUC within a $\approx1-4$ year period before declining in the longer term.

\subsection{Fairness Analysis}
\begin{table}[h]
\centering
\adjustbox{max width =.9\linewidth}{
\begin{tabular}{lllccccccccc}
\toprule
&& \multicolumn{5}{c}{\textbf{PH corpus}}& \multicolumn{5}{c}{\textbf{MIMIC-IV}}\\
\cmidrule(r){3-7}\cmidrule(r){8-12}
&&\textbf{G} & \textbf{\gnn}&\textbf{CLS}& \textbf{RVL}& \textbf{ft}&\textbf{G} &\textbf{\gnn}&\textbf{CLS} & \textbf{RVL} &\textbf{ft} \\
\midrule
\multirow{4}{*}{\rotatebox[origin=c]{90}{Ethnicity}}&\multirow{2}{*}{\rotatebox[origin=c]{90}{DPD}}&H&-0.12&-0.11&-0.05&	-0.02&&&&\\
&&N &0.12&0.11	&0.05&	0.02&&&&\\
\cmidrule{2-7}
&\multirow{2}{*}{\rotatebox[origin=c]{90}{EOD}}&H&	-0.08&-0.09	&0&	0.01&&&&\\
&&	N	&0.08&0.09	&0&	-0.01	\\
\midrule

\multirow{4}{*}{\rotatebox[origin=c]{90}{Gender}}&\multirow{2}{*}{\rotatebox[origin=c]{90}{DPD}}
&F&-0.04&-0.02&0.03&0.03& F &0.01&0.04&0.04&0.02\\
&&M&0.04&0.02&-0.03&-0.03& M &-0.01&-0.04&-0.04&-0.02\\
\cmidrule{2-12}
&\multirow{2}{*}{\rotatebox[origin=c]{90}{EOD}}
&F&0&0&0.06&0.06& F &0.12&0.2&-0.10&0.19\\
&&M&0&0&-0.06&-0.06&M&-0.12&-0.2&0.10&-0.19\\
\midrule
\multirow{8}{*}{\rotatebox[origin=c]{90}{Race}}&\multirow{4}{*}{\rotatebox[origin=c]{90}{DPD}}
&B&0.09&0.09&0.10&0.04& B &0.04&0.13&0.03&0.12\\
&&W&0&-0.01&-0.09&-0.03& W &0.04&-0.04&-0.03&-0.04\\
&&NI&-0.11&-0.10&-0.06&-0.03& H &-0.01&0.01&0.07&0.02\\
&&&&&&&A&-0.34&-0.32&-0.10&-0.31\\
\cmidrule{2-12}
&\multirow{4}{*}{\rotatebox[origin=c]{90}{EOD}}
&B&0.08&0.08&0.08&0.06& B &0.05&0.18&-0.06&0.16\\
&&W&0&0.03	&-0.09	&-0.03& W & 0.04&-0.06	&-0.01&	-0.05\\
&&NI&-0.11&-0.11&-0.02&-0.03& H &-0.03&-0.06&0.15&-0.05\\
&&&&&&& A &-0.37&-0.33&-0.03&-0.32\\
\bottomrule
\end{tabular}
}
\caption{DPD/EOD for subgroups (G): H (Hispanic), N (non-Hispanic), F (Female), M (Male), B (Black), W (White), NI (Unknown), A (Asian). Models: \gnn (\hitgnn), CLS (CLSTM), RVL(\reveal), and ft (Llama3.2-1B-ft).}
\label{tab:dem_metrics}
 \vspace{-5mm}
\end{table}
\begin{figure}[!htbp]
    \centering    \includegraphics[width=1\columnwidth, trim=5 25 0 0, clip]{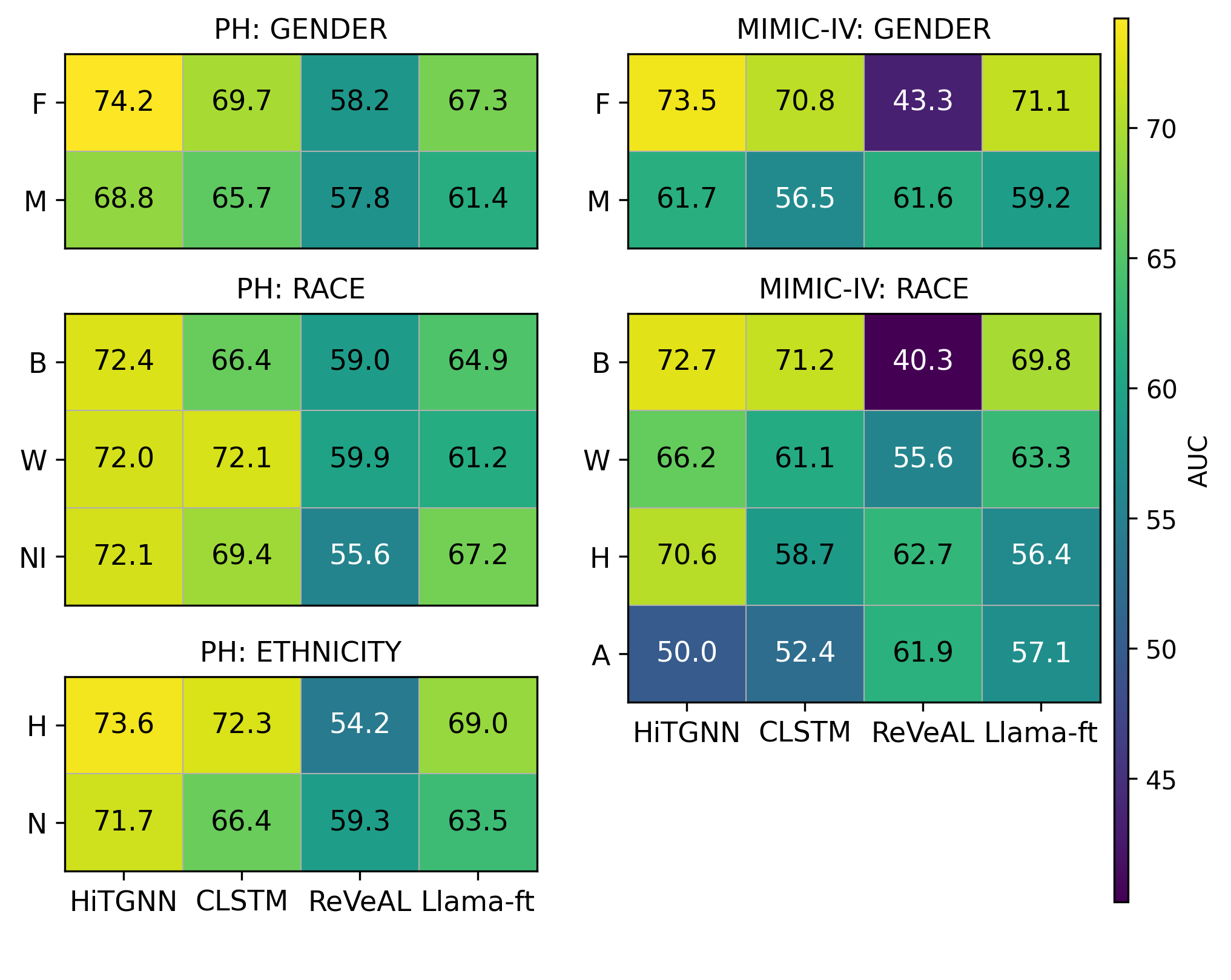}    
    \caption{AUC evaluated at subgroup level.}
    \label{fig:auc_subgroup}\vspace{-6mm}
\end{figure}
Table \ref{tab:dem_metrics} reveals consistent, corpus-dependent fairness trade-offs. In PH, all models exhibit negative DPD for Hispanic patients, with \hitgnn showing the largest imbalance (DPD:$-0.12$, EOD:$-0.08$). For gender, \hitgnn exhibits a mild imbalance against the Female majority (DPD $-0.04$; EOD $0$), while LLM-based models RVL (\reveal) and ft (Llama3.2-1B-ft) exhibit mild negative DPD and nontrivial EOD shifts ($-0.06$) against \textit{Males}. Race-related disparities are most pronounced for the minority Unknown (NI) group. In contrast, the majority Black patients are mildly favored across models. In MIMIC-IV, disparities against Males are mild in terms of DPD but substantial in terms of EOD, with RVL reversing direction to favor \textit{Male} patients. Among race groups, Asian patients experience the strongest negative shifts in both DPD and EOD across most models. RVL strongly favors \textit{Hispanics} and ft strongly favors \textit{Black} (B). Across both corpora, CLSTM is comparable to or more imbalanced than \hitgnn.

\hitgnn achieves the best AUC across subgroups and best worst-case subgroup protection in PH (Figure \ref{fig:auc_subgroup}). On MIMIC-IV, it again leads across groups for gender and race except for Asians, where it performs worst ($50\%$). Here, Llama-ft offers the best worst-case race protection $56.4\%$, highlighting group-specific trade-offs between aggregate discrimination and extreme subgroup harm.

\subsubsection{Demographic prior probing. }
Although \hitgnn does not use protected attributes, LLM-based models' overall performance improves when demographics are provided, raising the question of whether such gains reflect legitimate population-level risk factors or reliance on spurious demographic shortcuts learned during pretraining. To disentangle these effects, we conduct demographic prior probing analyses. Specifically, we construct deliberately underspecified prompts that provide only a single protected attribute (race, ethnicity, or gender), with no clinical evidence, to isolate reliance on demographic priors. Probes are run on Llama3.1-8B—the model used for explanation generation—using greedy decoding to obtain deterministic, highest-probability completions.
 
Population data show that diagnosed diabetes prevalence in the U.S. is highest among non-Hispanic Black adults, followed by Hispanic and non-Hispanic Asians, and lowest among non-Hispanic White adults \citep{CDC}, with higher risk in men than women \citep{kautzky2023sex}. Llama-8B sometimes reflects these patterns, assigning a higher risk to Black, Hispanic, and Asians relative to White patients. However, it also generates self-contradictory explanations, alternately asserting that ``as per CDC, Non-Hispanic whites have a higher prevalence of diagnosed diabetes compared to other racial and ethnic groups'', and that ``Type 2 Diabetes is higher among non-Hispanic blacks and Hispanics compared to non-Hispanic whites''. For gender, it generates similar contradictions, variously asserting ``women are at lower risk than men'' and ``men are generally at lower risk than women''. 

Appendix \ref{app:fairness} Table \ref{tab:dem-nodem-Llama} compares DPD and EOD with and without demographic conditioning, and finds that it affects fairness in corpus-dependent ways: in PH, it increases both the gender and racial imbalance. However, in MIMIC, it reduces gender imbalance but induces asymmetric racial shifts, substantially reducing bias toward Black patients while increasing it for White patients, and most strongly favoring Asian patients. 

Overall, \hitgnn offers more stable and higher average subgroup-level discrimination on both corpora. LLM-based models improve worst-case protection for race on MIMIC-IV. Analysis of LLMs with and without demographic conditioning shows their sensitivity to protected attributes can substantially reweight subgroup outcomes. Demographic prior probing reveals that LLM gains from demographic conditioning should be interpreted cautiously, as they may result from unstable group-level shortcuts rather than individual evidence. 

\subsection{Computation Time and Resources} \label{app:compute}
\begin{table}[!htbp]
\centering
\small
  \resizebox{\columnwidth}{!}{
\begin{tabular}{lccccc}
\toprule
\textbf{Model} & \textbf{Train-Time} & \textbf{Test-Time} & \textbf{Memory}&\#\textbf{Train Params} & \textbf{Input}\\
&(minutes) & (seconds)&(GB)&(Billion)&(\#notes)\\
\midrule
\hitgnn$^{\dagger}$ & 1.5  & 0.01  & 0.04 & 0.01 & 5  \\
CLSTM$^{\dagger}$ & 12 & 0.02  & 0.05& 0.05& 5  \\
Llama3.2-1B$^{Ex}$ & pre-trained & 6  & 2.4&\textsc{n/a} &2 \\
Llama3.1-8B$^{Ex}$  & pre-trained & 6  & 15&\textsc{n/a}&2  \\
Llama3.2-1B-ft &   18 hours  & 0.2 & 2.42 & 0.001 &2  \\
\reveal$^{Ex}$$^{\ddagger}$ & 30 hours & 62 & 17.42 &0.001 & 2\\
\bottomrule
\end{tabular}}
\caption{Computational efficiency of models. Inference time is per patient. $^{\dagger}$\hitgnn and CLSTM rely on MetaMap for UMLS linking. In this study, MetaMap was run sequentially due to API constraints, making the one-time preprocessing pipeline\textemdash extraction, linking, TimeGraph, and embeddings lookup\textemdash cost $\approx42$ seconds per PH note and $\approx15.6$ seconds per MIMIC-IV note. This step can be substantially accelerated via parallelization using recent tools such as ParallelPyMetaMap (\url{https://github.com/biomedicalinformaticsgroup/ParallelPyMetaMap}). $^{\ddagger}$The table excludes one-time data preparation costs for fine-tuning the \reveal model; the training time is over 5 reasoning paths, while inference time is over 10. $^{Ex}$These models generate explanations.}
\label{tab:compute}
\end{table}
Table \ref{tab:compute} summarizes the resource usage across models (excluding GPT-4o and DeepSeek-V3, which are accessed via API). A single A100 GPU with 80 Gb RAM is used for all experiments (accelerated inference is used for Llama experiments). Training the \hitgnn model takes $\approx1.5$ minutes, and inference per patient takes $\approx0.01$ seconds. For CLSTM, training the model takes $\approx12$ minutes, and inference on one patient takes $\approx0.02$ seconds. For inference with the Llama model, the average time required with explanations is approximately $\approx6$ seconds. Fine-tuning the model takes around 18 hours, and inference with this version takes 0.2 seconds. \reveal verifier training takes 30 hours (excluding training data preparation, which takes approximately 30 seconds per patient for obtaining explanations from Llama3.1-8B along 5 reasoning paths). Inference along 10 reasoning paths from this model takes $\approx62$ seconds (obtaining and verifying reasoning along 10 paths). The full Llama3.2-1B model requires $\approx2.4$ GB of memory, while Llama3.1-8B requires $\approx15$ GB. The LoRA adaptors require an additional 20 MB. In contrast, CLSTM takes 54 MB and \hitgnn is even lighter at 44 Mb, demonstrating impressive scalability in terms of both speed and memory usage. The \textsc{SpanTREx} model used by CLSTM and \hitgnn requires 0.44 GB of space.

\subsection{\hitgnn Ablations}

\subsubsection{Node Embeddings and Subgraph Variations:}
\begin{figure}[!htbp]
    \centering
    \begin{subfigure}[b]{0.5\linewidth}
        \centering
        \includegraphics[width=\linewidth, trim=20 6 12 0, clip]{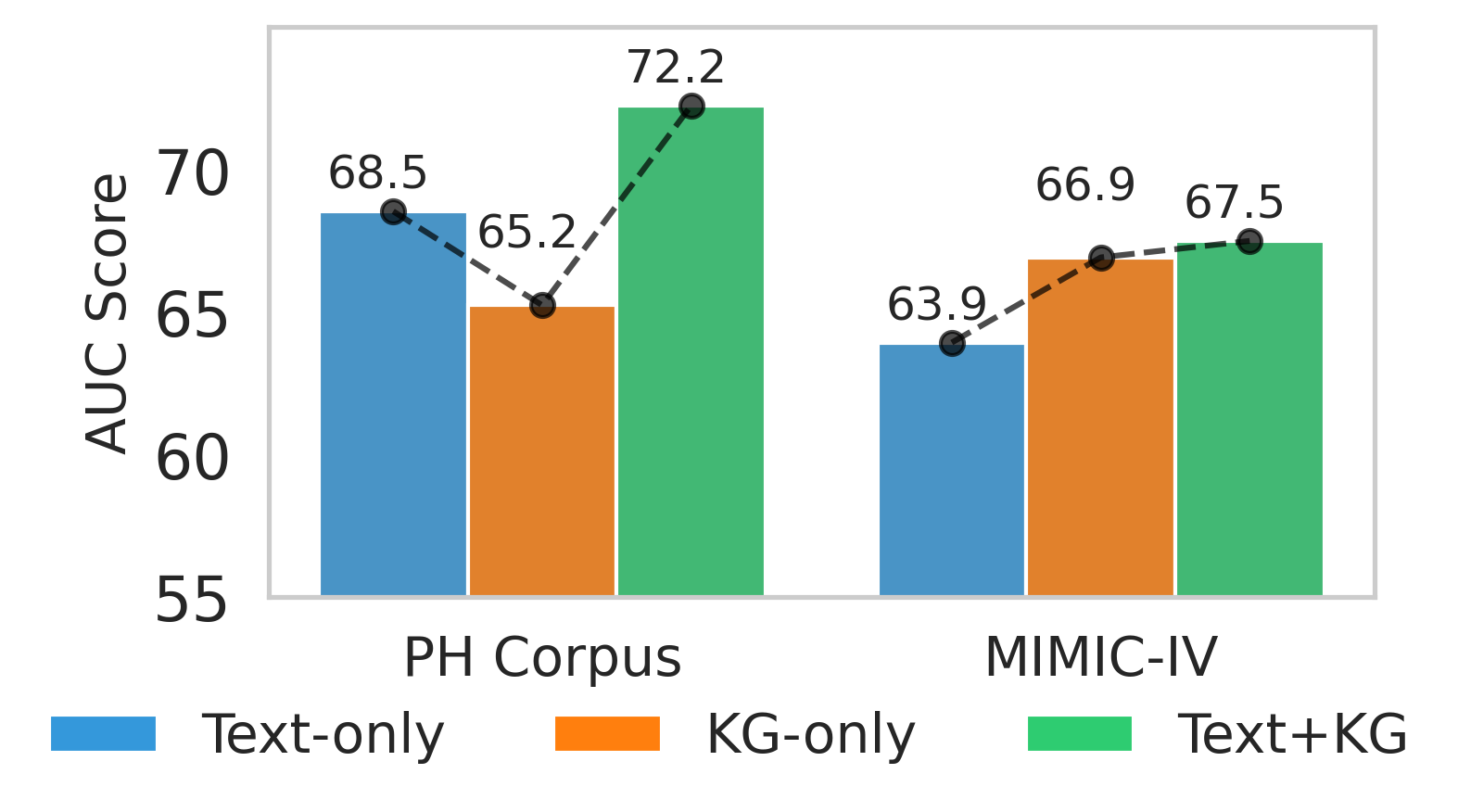}
        \caption{Embeddings}
        \label{fig:node_embedding_ablations}
        \end{subfigure}
    \hfill
    \begin{subfigure}[b]{0.47\linewidth}
        \centering
        \includegraphics[width=\linewidth, trim=25 5 5 0, clip]{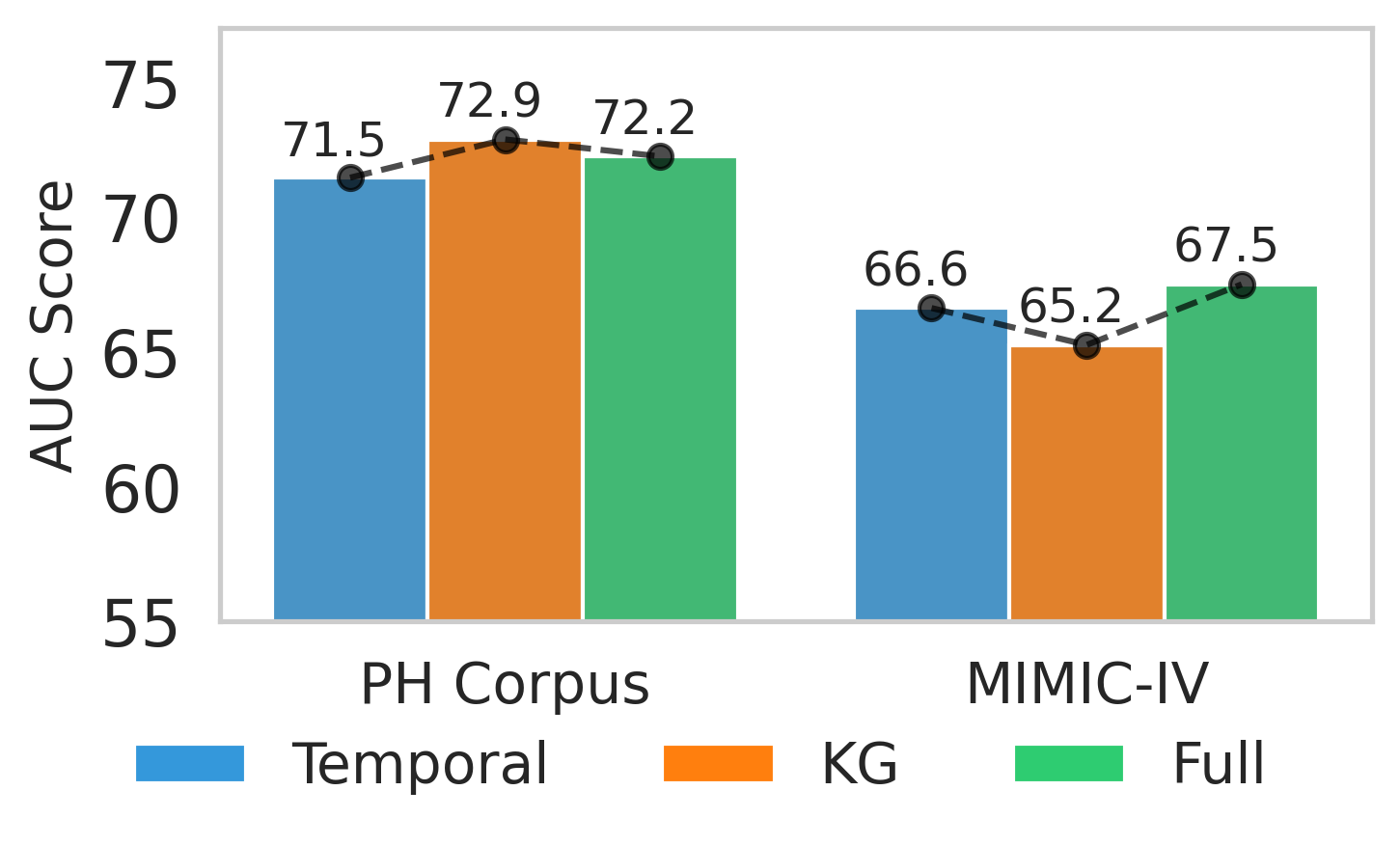}
        \caption{Subgraphs}
        \label{fig:subgraphsgnn}
    \end{subfigure}
    \caption{Node embedding and subgraph ablations.}
    \vspace{-5mm}
    \label{fig:hitgnn_ablations}
\end{figure}

Figure \ref{fig:hitgnn_ablations} compares \hitgnn performance across input choices based on AUC scores. Corresponding T2D recall trends are discussed in Appendix \ref{app:ablation_recall}. Figure \ref{fig:node_embedding_ablations} shows comparisons across node representations. Node initialization with in-context BioMedBERT embeddings (`Text-only') outperforms UMLS-CUI embeddings (`KG-only') on the PH corpus but underperforms it on MIMIC-IV. Combining both (`Text+KG') gives the best performance on PH but stays comparable to `KG-only' on MIMIC-IV (p-value=0.94).\footnote{P-values are reported based on bootstrap resampling test with 10,000 replicates.} Figure \ref{fig:subgraphsgnn} compares subgraphs: `Temporal' excludes the UMLS semantic network (KG), `KG' excludes temporal edges, and `Full' includes both. In PH, `KG' performs comparably to `Full' (p = 0.33). In MIMIC-IV, combining temporal edges and UMLS KG yields the best performance, but stays statistically comparable to `Temporal' alone (p = 0.35).

\subsubsection{Temporal Relation Extraction Models: }\label{sec:trexablation}
Table \ref{tab:end2end_ui} reports the performance of two recent state-of-the-art temporal relation extraction models, \textsc{GraphTREx} and \textsc{SpanTREx} \citep{chaturvedi-etal-2025-temporal}, trained on the I2B2-2012 corpus \citep{sun2013evaluating} and evaluated on a small sample of the PH corpus. This sample comprises nine notes jointly annotated by an NLP expert and a physician following the I2B2 entity annotation guidelines, using the NarrativeTime annotation tool \citep{rogers2024narrativetime}. This tool enables faster and denser annotations, resulting in $\approx280K$ relation pairs between $\approx2K$ entities. We report partial span matching $F_1$ scores for events and timex, and exact match $F_1$ for temporal relation (TLink). Consistent with prior work, \textsc{SpanTREx} achieves slightly higher $F_1$ scores for Event and TimeEx extraction, while \textsc{GraphTREx} yields higher precision. For TLink extraction, \textsc{GraphTREx} substantially outperforms \textsc{SpanTREx}, with approximately 14\% absolute gain in $F_1$. We also report details of the sample and the entity linking performance of Metamap (for clinical entities) and Recognizers-Text (for dates) in Appendix \ref{app:ph_annotation}. The results show high precision but lower recall, reflecting more fragmented clusters rather than noisy ones. 

Despite \textsc{GraphTREx} achieving higher scores in TLINK $F_1$ over \textsc{SpanTREx}, Table \ref{tab:hitgnn_trex_ablations} shows that downstream discrimination of \hitgnn stays comparable (on PH) or better (on MIMIC-IV) with \textsc{SpanTREx} over \textsc{GraphTREx}-extracted graphs. Still, the \textsc{GraphTREx}-based \hitgnn performs better (on PH) or comparable (on MIMIC-IV) to other baselines in Table \ref{tab:dp_results}. 
\vspace{-0.5mm}
\begin{table}[htbp]
    \centering
     \adjustbox{max width=1\columnwidth}{
    \begin{tabular}{l c c c}
\toprule
\textbf{Model}
 &\textbf{Event $F_1$} 
 &\textbf{Timex $F_1$}
 &\textbf{TLink $F_1$}\\
\midrule 
 \textsc{SpanTREx} &\textbf{74.39}&\textbf{69.46}&38.00\\
  \textsc{GraphTREx} &73.54&68.42&\textbf{51.97}\\
\bottomrule
\end{tabular}
}
\caption{Performance of state-of-the-art Temporal Relation Extraction models on PH subsample.}
    \label{tab:end2end_ui}
    \vspace{-6mm}
\end{table}
\begin{table}[h]      
    \centering
    \adjustbox{max width=.7\columnwidth}{
    \begin{tabular}{l c c}
    \toprule
    \multirow{2}{*}{\textbf{\trex Model}}&\multicolumn{2}{c}{\textbf{\hitgnn AUC}}\\
    \cmidrule{2-3}
    &\textbf{PH} &\textbf{MIMIC-IV}\\
        \midrule           \textsc{SpanTREx}&\textbf{72.24}&\textbf{67.49} \\
        \textsc{GraphTREx}& 71.81& 64.27\\
        \bottomrule
    \end{tabular}}
    \caption{\hitgnn AUC scores using \textsc{SpanTREx} vs. \textsc{GraphTREx} extracted temporal graphs.}    
    \label{tab:hitgnn_trex_ablations}
     \vspace{-4mm}
\end{table}

\subsubsection{Pipeline Component Ablations: }
Table \ref{tab:hitgnn_pipeline_ablations} presents an ablation study of the \hitgnn pipeline, examining the contribution of entity linking, TimeGraph-based filtering, and flipping \textit{After} edges to their inverse \textit{Before} across corpora. For the PH corpus, the full pipeline (linking + TimeGraph + flipping) achieves the highest AUC (72.24), indicating that jointly leveraging external knowledge, enforcing temporal consistency, and temporal relation simplification improves the overall discrimination with temporal edges. In contrast, for MIMIC-IV, the highest AUC (67.49) is achieved using entity linking alone, while normalization via TimeGraph construction or label flipping does not consistently improve AUC. This contrast is consistent with the dataset characteristics discussed in Section \ref{sec:data}, Table \ref{tab:graph-stats}: MIMIC-IV has sparser event–temporal graphs, whereas the PH corpus contains denser graphs extracted from longer notes. In PH, TimeGraph and label flipping help mitigate noise in raw temporal relations, but in MIMIC, they may sparsify already weak but useful temporal signals.

\begin{table}[h]      
    \centering
    \adjustbox{max width=1\columnwidth}{
    \begin{tabular}{ccc cc cc}
    \toprule
\multicolumn{3}{c}{\textbf{Pipeline Configuration}}
&\multirow{2}{*}{\textbf{PH Corpus}}&\multirow{2}{*}{\textbf{MIMIC-IV}}\\
\cmidrule{0-2}%
 \textbf{Linking} & \textbf{TimeGraph} & \textbf{Flipping}%
 \\
        \midrule   
        
        \checkmark&\checkmark&\checkmark &\textbf{72.24}
        & 64.00
        \\
        \ding{55}&\ding{55}&\ding{55}& 65.28
        & 63.20  
        \\
        \ding{55}&\checkmark&\ding{55} & 62.46 
        &63.05 
        \\ 
        \ding{55}&\checkmark&\checkmark& 65.51 
        &64.80 
        \\ 
        \checkmark&\ding{55}&\ding{55}&64.84 
        &\textbf{67.49}
        \\
        \checkmark&\ding{55}&\checkmark& 64.98  
        & 67.05          \\
        \bottomrule
    \end{tabular}}
    \caption{\hitgnn AUC variations with pipeline.}    
    \label{tab:hitgnn_pipeline_ablations}
    \vspace{-5mm}
\end{table}

\subsubsection{\gnn Architecture Ablations: } Table \ref{tab:hitgnn_gnn_ablations} shows variations with \gnn encoders.\footnote{These configurations use mean pooling. With max pooling, performance drops, e.g., GraphSAGE AUC decreases to 68.02 on PH and 60.45 on MIMIC-IV.} GraphSAGE has the strongest performance across corpora. More expressive encoders (GAT, RGCN, HGTConv) do not provide further gains. HGT (Heterogeneous Graph Transformer) performs weakest, likely because its type-specific message passing and type-wise pooling introduce substantial sparsity and imbalance.\footnote{HGT parameterizes interactions over (source node type, relation type, target node type) triplets, leading to a combinatorial increase in edge types; due to computational constraints and to limit further sparsity, we exclude UMLS KG.} This suggests that \hitgnn benefits more from stable neighborhood aggregation than highly parameterized models.
 
\begin{table}[htbp]      
    \centering
    \adjustbox{max width=.75\columnwidth}{
    \begin{tabular}{l ccc ccc}
    \toprule
\textbf{GNN Model}&\textbf{PH Corpus}&\textbf{MIMIC-IV}\\
        \midrule   
        
        \textbf{GraphSAGE} &\textbf{72.24} 
        &\textbf{67.49}
        \\
        \textbf{RGCN} &66.09
        & 65.75 
        \\
        \textbf{GAT} &65.99
        &64.73
        \\
        \textbf{HGT}&54.67&59.62        \\  
        \bottomrule
    \end{tabular}}
    \caption{AUC across \gnn variations in \hitgnn.}    
    \label{tab:hitgnn_gnn_ablations}
    \vspace{-5mm}
\end{table}
\subsection{Clinical Analysis of \reveal Explanations.}
To better understand how \reveal reasoning aligns with contemporaneous clinical reasoning, we perform a qualitative analysis of model explanations on a small set of cases from the PH test sample. A senior endocrinologist independently assessed risk based on the last two clinical notes and then reassessed after reviewing the \reveal explanations over the same inputs. In all cases, explanations did not alter the clinician's original judgment, suggesting that they do not exert undue persuasive influence. In one case, the expert supports the model's judgment, in contradiction with the ground truth, possibly due to delayed onset or left censoring. These explanation-expert assessments are provided in Appendix \ref{app:human-eval}. They illustrate both cases where the model's explanations closely mirror expert judgment and cases where plausible-sounding explanations over-emphasize weak or irrelevant signals, highlighting important limitations of current LLMs in challenging clinical scenarios. 

\section{Conclusion}
We address the challenge of clinical risk prediction from longitudinal notes by moving beyond static snapshots to dynamic patient representations. To this end, we present \hitgnn, a temporally grounded event-centric \gnn augmented with clinical KGs. \hitgnn attains superior predictive performance, especially on the immediate-risk horizon, while remaining computationally efficient. Complementing \hitgnn, we introduce \reveal, a test-time LLM scaling method for interpretable predictions. By utilizing a smaller LLM to verify rationales from a larger LLM, \reveal offers a promising balance between performance and explainability, while much larger models like GPT-4o exhibit limited zero-shot performance on this complex clinical task. By constructing temporally realistic, demographically matched cohorts, we avoid post-diagnosis data leakage and enable controlled fairness evaluation. Our fairness analysis finds that even without demographic conditioning, \hitgnn demonstrates more stable and higher subgroup-level discrimination across datasets and groups, except for the Asian minorities. On the other hand, LLM assessments are sensitive to protected attributes and may rely on unstable demographic shortcuts.

Ablations examine the role of multiple information sources in \hitgnn: enriching node representations with both contextualized and UMLS KG embeddings improves performance on PH but remains comparable to KG-only initialization on MIMIC-IV, temporal edges do not impact AUC but improve T2D recall on PH, while on MIMIC-IV they improve AUC but reduce recall. PH benefits from a full pipeline, whereas MIMIC-IV performs best with KG-linking but avoiding temporal edge pruning. We further find that \hitgnn is robust to variation in upstream \trex quality, remaining better or comparable to other baselines. Overall \hitgnn's downstream T2D performance reflects interactions among multiple upstream signals and corpus-specific characteristics.

While we currently use a hierarchical approach to aggregate temporal graphs within and across visits, the lack of publicly available cross-document temporal and coreference corpora limits explicit supervision to link events across notes; future works in this direction could enable better-connected patient timelines. \hitgnn's modular design also enables component-level optimization\textemdash such as adopting newer biomedical linking tools \citep{KRALJEVIC2021102083} and embeddings from larger language models. Improving subgroup fairness for underrepresented populations is another important direction for future work. For example, in MIMIC-IV, lower performance for the Asian subgroup may reflect both limited sample size and potential documentation bias; future work should explore cross-institutional data and targeted augmentation to support more equitable performance. Additional directions include developing alignment between graph-based reasoning and LLM explanations and extending these methods to other chronic diseases. More broadly, the proposed framework may also generalize to other domains requiring fine-grained temporal reasoning over longitudinal textual data, including financial forecasting and conflict prediction from news streams.
\section*{Ethics Statement}
The use of the PH corpus was approved by the institutional review board (IRB) at the data hosting institution, and MIMIC-IV was reviewed by the IRB at Beth Israel Deaconess Medical Center. Both datasets were obtained after completing the required training and accessed under secure protocols in accordance with institutional and data use policies \citep{PhysioNet-MIMICiv-1.0}. We use the Azure OpenAI Service for GPT and Deepseek API hosted by togetherai for LLM inference on MIMIC-IV and opted out of human review due to restrictions on third-party processing of sensitive clinical data (\url{https://physionet.org/news/post/gpt-responsible-use/}). 

The models presented in this work are intended as clinician-in-the-loop risk stratification or early-warning tools and should not be used as standalone diagnostic and medical decision-making tools without additional clinical validation.

\section*{Acknowledgments}
The authors thank Sourav Medya for helpful feedback and Luke Lauridsen for assistance in preparing the annotated sample used in our additional analysis (Section \ref{sec:trexablation}). We also thank action editor Yulan He and reviewers for their helpful comments and suggestions. This work is also partially supported by the NSF under award IIS-2312862.

\bibliography{tacl2021v1}
\bibliographystyle{acl_natbib}
\clearpage
\appendix
\section{Data Curation Details}
\label{app:data}
\subsection{PH corpus}
We show the data-preparation steps in Table \ref{tab:data_filter_final_ui}. The initial data is 164,910 patients, 15,140 of whom were diagnosed with Type 2 Diabetes, and 3,708,168 clinical notes. We exclude all post-diagnosis notes (these include notes after 3 days pre-diagnosis date). The 3-day pre-diagnosis window is considered to account for possible lag in transforming information from notes to ICD codes in structured tables. This filter ensures we only consider pre-diagnosis information, leaving us with around 3 million notes. We then exclude the rare demographic groups. These include race groups Native Hawaiian/Other Pacific Islander/American Indian/Alaska Natives/Asian, and unknown Gender/Ethnicity. We filter the data by note type and author, as it includes all providers and note types, some of which are uninformative. This subset of notes is determined with the help of a senior expert endocrinologist and is provided below:
\paragraph{Important Note Types}
 Inpatient Progress Note, Progress Notes, Family Medicine Note, History and Physical Note, Emergency Department Note, Discharge Summary, Nutrition Note, Endocrinology Note, Specialty Pharmacy Services, General Eye Note, Diet Instructions, ED Notes, Outpatient Pharmacy Note, Discharge Note, Dialysis Rounding, Diabetes Education.

 \paragraph{Important Author Types}
 Unspecified, Anesthesiologist, Care Coordinator, Dentist, Dietician Student, Fellow, Licensed Clinical Social Worker, Licensed Practical Nurse, Medical Assistant, Medical Student, Mental Health Counselor, Midwife, Nurse Practitioner, Nursing Student, Occupational Therapist, Occupational Therapy Assistant, Occupational Therapy Student, Optometrist, Oral Surgeon, Pharmacist, Physical Therapist, Physical Therapy Assistant, Physical Therapy Student, Physician, Physician Assistant, Registered Dietitian.

We then remove notes shorter than 100 words, following previous work in health-risk prediction \citep{ghassemi2015multivariate}. Upon manual inspection of some of the notes in this filtered subset from the pre-diagnosis date, we find that a diagnosis of T2D is already present but not entered in the structured records, or it is entered much later. As discussed previously, such ICD-coding errors are common \citep{hersh2013caveats}. This motivated us to apply an additional filter using a popular large language model (LLM), Llama3.1-8B. LLMs are exceptionally good at extracting explicit information already present in the text. We apply two different prompts (these are variations of the prompt in Figure \ref{fig:Llama_cleaning_prompt}), asking the model to identify if an explicit diagnosis is already present in the note of the T2D cohort (due to scalability challenges, we do not apply the LLM filter to the NoD cohort). Manual inspection of 50 samples reveals higher accuracy if the model answers yes to both prompts. We readjust the earliest diagnosis date accordingly and drop all notes on or after that date. This further reduces our T2D subset to 1717 patients. We split this data into a training-testing split in an 80:20 ratio, stratified by the diagnosis group.

\begin{table}[htbp]
    \centering
    \begin{adjustbox}{max width=\linewidth}
    \begin{tabular}{l|c|c|c}
    \hline
   Filtering Step & \textbf{\#NoD}&\textbf{\#T2D}&\textbf{\#Notes}\\
   \hline
   Initial&164,910&15,140 &3,708,168\\
   \hline
   Pre-diagnosis notes&164,910&4,138& 3,146,140\\
Non-rare demographics&149,673& 3,878&2,909,417\\
Important Note and Author-types &108,635 &2,369 &1,451,626\\
Note length& 106,222& 2,302&1,050,839\\
LLM for diagnosis date adjustment  &106,222 &1,717&1,039,043\\
\hline
\end{tabular}
\end{adjustbox}
    \caption{Data filtering steps and number of non-Diabetic (NoD) and T2D patients, and notes after each step for the PH corpus.}
    \label{tab:data_filter_final_ui}
    \vspace{-3mm}
\end{table}
We then apply propensity-weighted matching to construct a treatment (T2D) and control (NoD) matched test sample \citep{rosenbaum1985constructing}. We use the demographic attributes to match the covariates. For this, we first train a logistic regression model to estimate propensity scores (likelihood of being in the T2D group, given the demographic attributes) (the model achieves an AUC score of 73.40\%). We use a greedy nearest-neighbor 1:1 matching without replacement to construct the final control group (NoD). For each treated sample (T2D), we find an unmatched control (NoD) with the closest propensity score, ensuring each control is used only once. Table \ref{tab:covariate_balance} presents the covariate balance before and after the matching.

\begin{table}[ht]
\centering
\begin{adjustbox}{max width=.8\columnwidth}
\begin{tabular}{lcc}
\toprule
\textbf{Covariate} & \textbf{Before} & \textbf{After} \\
\midrule
AGE      & 0.683 & 0.004 \\
Binarized GENDER (M=1)       & 0.028 & 0 \\
Binary RACE NI         & -0.057 &0 \\
Binarized RACE W           & -0.270 & 0.005 \\
Binarized ETHNICITY (Y=1)     & 0.002 &0 \\
\bottomrule
\end{tabular}
\end{adjustbox}
\caption{Covariate balance using standardized mean differences (SMD) before and after propensity score matching for PH corpus test set.}
\label{tab:covariate_balance}
\vspace{-5mm}
\end{table}
\subsection{MIMIC-IV T2D Data}
We select the treatment group as the subset of patients with ICD code E11 (Type 2 Diabetes). We exclude patients with ICD codes from related diseases (E08-E13, O24) to exclude other types of diabetes. We coarsify race as Hispanic if Race is Hispanic/Latino, BLACK/AFRICAN AMERICAN if race is either Black/African, and drop patients with race group listed as `other'. Thereafter, we only retain pre-diagnosis notes. MIMIC-IV notes contain discharge summaries from which we retain ``Chief Complaint'', ``History of Present Illness'', and ``Discharge Instructions'' sections and drop all patients with more than 20 admissions following earlier works. We then drop duplicated notes for a patient, and notes that are too short (<.05 quantile) and too long (>0.95 quantile). Finally, we apply an LLM filter on the T2D subgroup to correct the diagnosis date. We use both Llama3.1-8B and GPT-4o and manual cleaning to construct a robust subset. Finally, we apply propensity-weighted matched sampling to get a balanced control group. The data filtering steps are detailed in Table \ref{tab:data_filter_final_MIMIC}. The AUC score of the logistic regression model to estimate the propensity score is 63.64\%. The covariate balance before and after matching is presented in Table \ref{tab:covariate_balance_MIMIC}. The data is split into training and test in a 90:10 ratio.
\begin{table}[htbp]
    \centering
    \begin{adjustbox}{max width=\linewidth}
    \begin{tabular}{l|c|c|c}
    \hline
   Filtering Step & \textbf{\#Notes}& \textbf{\#NoD}&\textbf{\#T2D}\\
   \hline
Initial Data &4,756,326& \multicolumn{2}{c}{180,640}\\
No other types of diabetes &4,578,799&\multicolumn{2}{c}{178,524}\\
race not `other' &4,033,399&\multicolumn{2}{c}{156,513}\\
\hline
Assign cohort &4,033,399&142,412&14,101\\
pre-diagnosis notes&3,220,793&142,412&5,952\\
Drop if more than 20 admissions&235,727&59,566& 5,847\\
No Duplicates&184,421&54,803& 5,272\\
Notes length in .05-.95 quantile&175,201&54,170&5,207\\
No duplicates across sections& 174,995&54,170&5,207\\
    \hline
    LLM + matched controls Filter &14,957&2,909&2,909\\
    \hline
\end{tabular}
\end{adjustbox}
     \caption{Data filtering steps and number of non-Diabetic (NoD) and T2D patients, and notes after each step for MIMIC-IV.}    \label{tab:data_filter_final_MIMIC}
     \vspace{-3mm}
\end{table}

\begin{table}[ht]
\centering
\begin{adjustbox}{max width=.8\columnwidth}
\begin{tabular}{lcc}
\toprule
\textbf{Covariate} & \textbf{Before} & \textbf{After} \\
\midrule
AGE      & 0.29& 0.001 \\
Binarized GENDER  (M)      & 0.12& -0.001 \\
Binary RACE (B)         &  0.316 & 0 \\
Binarized RACE (H)          & 0.110 & 0.001\\
Binarized RACE (W) &-0.334 & 0\\
\bottomrule
\end{tabular}
\end{adjustbox}
\caption{Covariate balance using standardized mean differences (SMD) before and after propensity score matching on MIMIC-IV test set.}
\label{tab:covariate_balance_MIMIC}
\vspace{-3mm}
\end{table}

\section{Implementation Details}
\label{app:implementation}
\subsection{Prompts}
Figure \ref{fig:Llama_cleaning_prompt} presents the prompt used to identify a known diagnosis of T2D from a clinical note. Prompts for LLM-based experiments are in Figures \ref{fig:Llama_dp_reasoner_prompt}\textendash\ref{fig:reveal_prompts}.
\begin{figure}
\centering
\noindent
\begin{tcolorbox}[title=Prompt]
\small
You are a helpful assistant who can read and identify required information from given clinical notes, either mentioned explicitly or in ICD code format.

\textbf{Required information:} a diagnosis of \textbf{Diabetes} for the given patient.

Please reply \texttt{True} if an explicit diagnosis is mentioned in the note.

Please answer \texttt{False} if:
\begin{itemize}
  \item there is no diagnosis,
  \item the note negates the diagnosis (e.g., ``no diagnosis of diabetes'' or ``diabetes: negative''),
  \item if the diagnosis is associated with someone other than the patient (e.g., family).
\end{itemize}

Do not output anything other than \texttt{True}/\texttt{False}.

\textbf{NOTE:} \texttt{[NOTE\_TEXT]}

\textbf{Output:}
\end{tcolorbox}
\caption{Identifying known T2D in a note.}
\label{fig:Llama_cleaning_prompt}
\vspace{-3mm}
\end{figure}
\begin{figure}[htbp]
\centering
\begin{tcolorbox}[title=System Prompt: Reasoner, colback=gray!5, colframe=black!50, width=\linewidth]
\small
You are a helpful medical assistant. Based on the patient's information and the history of medical notes from one or more visits, assess if this patient has a likelihood of being diagnosed with Type 2 Diabetes in the near future.\\
Make the assessment based on \textbf{concrete} medical evidence and how the patient's condition has progressed. Do not solely rely on age as a risk factor.\\[0.5em]
\textbf{\# Prediction Format:}\\
\textbf{\#\# Risk of Type 2 Diabetes:} **[True/False]\\
\textbf{\#\# Explanation:} [Provide a brief explanation of the reasoning for the prediction.]
\end{tcolorbox}
\vspace{0.5mm}
\begin{tcolorbox}[title=User Prompt, 
colback=gray!5, colframe=black!50, width=\linewidth,
]
\small
Case History:\\[0.5em]
Patient is \texttt{\{age\}} y.o. \texttt{\{race\}} \texttt{\{ethnicity\}} \texttt{\{gender\}}. \\[0.5em]
\texttt{Note Date: \{date1\}} \\
\texttt{\{note-type1\} note} \\
\texttt{\{note-text1\}} \\[0.5em]
\texttt{Note Date: \{date2\}} \\
\texttt{\{note-type2\} note} \\
\texttt{\{note-text2\}}
\end{tcolorbox}

\begin{tcolorbox}[
  colback=blue!5,
  colframe=blue!20,
  boxrule=0.1mm,
  width=\linewidth,
  sharp corners,
  before skip=0pt,
  after skip=0pt
]
\footnotesize
\textit{Note: \{\texttt{ethnicity}\} and \{\texttt{note-type}\} are excluded for the MIMIC-IV. \{\texttt{ethnicity}\} is also excluded for the PH corpus if the value is `NI'.}
\end{tcolorbox}
\caption{T2D risk from a reasoning model.}
\label{fig:Llama_dp_reasoner_prompt}
\vspace{-8mm}
\end{figure}

\begin{figure}[htbp]
\centering
\begin{tcolorbox}[title=System Prompt: Fine-tuned Reasoner, colback=gray!5, colframe=black!50, width=\linewidth]
\small
You are a helpful medical assistant. Based on the patient's information and the history of medical notes from one or more visits, assess if this patient has a likelihood of being diagnosed with Type 2 Diabetes in the near future. \\
You must respond with either:
\begin{itemize}
    \item \texttt{true}
    \item \texttt{false}
\end{itemize}
No explanation is needed. Only output one of the two labels above.
\end{tcolorbox}
\caption{T2D risk Prediction without explanation. User prompt is the same as Figure \ref{fig:Llama_dp_reasoner_prompt}.}
\label{fig:Llama_dp_finetune_prompt}
\vspace{-5mm}
\end{figure}

\begin{figure}
\centering
\begin{tcolorbox}[title=System Prompt: Fine-tuned Verifier, colback=gray!5, colframe=black!50, width=\linewidth ]
\small
You are a medical reasoning assistant. Given a sequence of clinical notes from a patient and an analysis predicting the risk of an imminent Type 2 Diabetes (T2D) diagnosis, judge if the analysis is correct or incorrect.\\
You must respond with either:
\begin{itemize}
    \item \texttt{correct}
    \item \texttt{incorrect}
\end{itemize}

No explanation is needed. Only output one of the two labels above.
\end{tcolorbox}
\vspace{-0.5em}
\begin{tcolorbox}[title=User Prompt, 
colback=gray!5, colframe=black!50, width=\linewidth,
]
\small
Case History:\\[0.5em]
Patient is \texttt{\{age\}} y.o. \texttt{\{race\}} \texttt{\{ethnicity\}} \texttt{\{gender\}}. \\[0.5em]
\texttt{Note Date: \{date1\}} \\
\texttt{\{note-type1\} note} \\
\texttt{\{note-text1\}} \\[0.5em]
\texttt{Note Date: \{date2\}} \\
\texttt{\{note-type2\} note} \\
\texttt{\{note-text2\}}\\[0.5em]
\texttt{Analysis}\\
\texttt{\{Reasoner Output\}}
\end{tcolorbox}
\caption{Fine-tuning the verifier model in the \reveal to verify if the T2D prediction assessment was correct based on model input and output.}
\label{fig:reveal_prompts}
\vspace{-5mm}
\end{figure}

\subsection{Experimental Settings}
For supervised learning/fine-tuning experiments, a separate model is trained for each dataset. 
\paragraph{GNN}
For GNN (and CLSTM) experiments, we use graphs from the last five notes (we experimented with 1,2,5,10). We train the models using 3-fold stratified cross-validation on the training set and apply early stopping based on fold-specific validation performance, saving the best model for each fold. Hyperparameters and \gnn architectural choices are selected based on mean validation performance across folds and fixed before evaluation on the held-out test set. \texttt{SAGEConv} operator from PyTorch Geometric with default settings is used. We experiment with $1-5$ \gnn layers and find the best performance with $k=2$ layers for PH and $k=1$ for MIMIC-IV. For aggregating the nodes for graph representation, we also experiment with max pooling and pooling only the nodes corresponding to document creation time (DCT/anchor nodes); neither works well. In place of a BiLSTM to model multi-document graphs, we also test LSTM, simple attention-based aggregation of cross-visit graph representations, and transformer-based aggregation, but all underperform compared to BiLSTM. 

\paragraph{LLM}
Due to computational constraints, we only fine-tune a smaller Llama3.2-1B model instead of the larger Llama3.1-8B variant, which exhausts memory even with a batch size of 1 on an A100 80GB GPU. For \reveal, we use Llama3.1-8B as the zero-shot reasoner and Llama3.2-1B as the fine-tuned verifier. We experimented with 1, 2, 3, 5, 8, and 10 notes and found that including the last two notes yields the best performance. As the notes are long and detailed, experiments with additional notes cause the LLM to underperform. In contrast, as GNN uses a structured summary of the notes in the form of temporal graphs, it can incorporate additional information effectively and gives the best performance with five notes. We also observe a slight improvement in LLM accuracy when demographic details are included, while it reduces the performance of CLSTM and \hitgnn. 

All LLM fine-tuning experiments are conducted on  Llama3.2-1B using a batch size of 1 due to the length of input sequences and memory limitations. For each dataset,  5\% random held-out validation set is reserved for hyperparameter tuning, stratified by the T2D/NoD outcome. We train the model for 30 epochs using the AdamW optimizer with a learning rate of $6e^{-06}$ and a weight decay of 0.01.\footnote{We experimented with learning rates of $5e^{-05}$ and $1e^{-06}$ that had lower performance.} A linear learning rate schedule without warm-up is applied. Gradient accumulation is used with a factor of 2 to stabilize updates, given the small batch size. The loss function is standard cross-entropy with mean reduction. In our setup, LoRA is applied to the query and value projection layers (q\_proj and v\_proj) within the transformer blocks. We use a rank of 8, a LoRA scaling factor (alpha) of 16, and a dropout rate of 0.1 during training. This configuration results in only a small fraction of the model parameters being updated ($850K/1.24B\approx0.07\%$), making the training process efficient (31 minutes per epoch for PH corpus and 32 for MIMIC-IV) and tractable on a single A100 80GB GPU while still allowing the model to adapt to the binary classification task.

In the fine-tuning experiments, the pretrained Llama3.2-1B model initially performs near chance level, with a balanced accuracy of 54.50\% on the validation set and 0.8\% of outputs falling outside the valid label set. After fine-tuning, the best model achieves an accuracy of 70.50\% on the validation set with no invalid predictions. On MIMIC-IV, the initial validation set accuracy is much lower at 44.00\%, with 18.1\% of predictions being invalid. This improves to 62.50\% and zero invalid predictions after fine-tuning. 

For proper supervision for \reveal, we restrict training to cases where the 8B reasoner produces both `true' and `false' predictions across the five runs. This ensures that each case includes both valid and invalid reasoning paths, enabling the verifier to learn discriminative patterns. We have $5{,}550$ training examples for the PH corpus and $9{,}180$ for MIMIC-IV. A 5\% random held-out validation set stratified by the `correct/incorrect' label is reserved from each. After each epoch, the model is evaluated on the validation set, and the best-performing model checkpoint is saved. The verifier 1B model starts with a base accuracy of 43.2\% and 1\% invalid predictions on the PH corpus, and the accuracy improves to 84.9\% and no invalid predictions in 30 epochs. For MIMIC-IV, the accuracy improves from 58.8\% initially to 67.5\%. 

\section{Entity, Temporal relation, Coreference Annotation Details}
\label{app:ph_annotation}
\begin{table}[htbp]
\centering
\small
\begin{minipage}[t]{0.48\columnwidth}
\centering
\adjustbox{max width=\columnwidth}{
\begin{tabular}{lc}
\toprule
\textbf{Entity Type} & \textbf{Freq.} \\
\midrule
PROBLEM & 373 \\
TREATMENT & 120 \\
TEST & 325 \\
OCCURRENCE & 489 \\
CLINICAL\_DEPT & 41 \\
EVIDENTIAL & 90 \\
DATE & 214 \\
TIME & 64 \\
DURATION & 67 \\
FREQUENCY & 74 \\
\bottomrule
\end{tabular}}
\end{minipage}
\hfill
\begin{minipage}[t]{0.48\columnwidth}
\centering
\adjustbox{max width=\columnwidth}{
\begin{tabular}{lc}
\toprule
\textbf{Relation Type} & \textbf{Freq.} \\
\midrule
Before & 83{,}392 \\
After & 83{,}392 \\
Simultaneous & 112{,}776 \\
\bottomrule
\end{tabular}}
\end{minipage}
\caption{Annotation details for 9 PH corpus notes comprising 10{,}936 tokens.}
\label{tab:phsample_entity}
\vspace{-5mm}
\end{table}
\begin{table}[h]
\centering
\begin{adjustbox}{max width=.7\columnwidth}
    \begin{tabular}{lccc}
\toprule
\textbf{Entity Type} & \textbf{P} & \textbf{R} & $\mathbf{F_1}$ \\
\midrule
PROBLEM          & 78.2 & 37.2 & 48.3 \\
TREATMENT        & 54.6 & 48.3 & 50.7 \\
TEST             & 47.0 & 63.3 & 48.6 \\
CLINICAL\_DEPT   & 29.2 & 26.5 & 27.5 \\
DATE             & 98.7 & 67.2 & 77.0 \\
\midrule
Selected 5 Types (Micro)  &77.2 &52.9 &61.9\\
\midrule
All 10 Types (Micro)&77.2 & 40.2 & 52.2\\
\bottomrule
\end{tabular}
\end{adjustbox}
\caption{Coreference resolution performance.}
\label{tab:coref}
\vspace{-5mm}
\end{table}
\begin{table*}
\centering
    \adjustbox{max width=\linewidth}{
    \begin{tabular}{l|cc|cc|cc|cc|cc|cc}
    \toprule
    \multirow{3}{*}{Model Pair}&\multicolumn{6}{c|}{\textbf{PH Corpus}}&\multicolumn{6}{c}{\textbf{MIMIC-IV}}\\
    \cline{2-13}
    &\textbf{AUC}&$\mathbf{F_1}$&\multicolumn{2}{c|}{\textbf{T2D}}&\multicolumn{2}{c|}{\textbf{NoD}}&\textbf{AUC}&$\mathbf{F_1}$&\multicolumn{2}{c|}{\textbf{T2D}}&\multicolumn{2}{c}{\textbf{NoD}}\\
        \cline{4-7}\cline{10-13}
         &  &&P & R &P & R& &&P & R&P & R \\
        \midrule   
ReVeAL vs. Llama3.2-1B-ft
& -0.07* & 0.04  & 0.07*** & -0.18*** & -0.05* & 0.23*** 
& -0.11* & -0.08* & -0.09 & -0.11* & -0.06* & -0.03 \\
&(0.03)&(0.02)&(0.02) & (0.03)& (0.03)& (0.03)& (0.05) & (0.04) & (0.06) & (0.06) & (0.03) & (0.05) \\
CLSTM vs Llama3.2-1B-ft&0.03&0.04 &0.04**& -0.02 & 0.02&0.09**& -0.01&0 &0& 0.01&0 &-0.01 \\ 
&(0.02)&(0.02)&(0.02)&(0.03)&(0.03)&(0.03)&(0.03)&(0.01)&(0.02)&(0.01)&(0.01)&(0.02)\\
CLSTM vs ReVeAL&0.10***& 0&-0.04 & 0.16***&0.06**&-0.14***& 0.10 &0.08 &0.09&0.12*&0.06 &0.02\\
&(0.03) & (0.02)& (0.02)&(0.03)&(0.03)&(0.03)&(0.05)&(0.04)&(0.06)&(0.06)&(0.03)& (0.05)\\

ReVeAL vs. \hitgnn
& -0.14*** & -0.02 & 0.02  & -0.16*** & -0.07** & 0.11***
& -0.14** & -0.10** & -0.02& -0.16** & -0.11* & -0.16**\\
& (0.03) &(0.02) & (0.02) & (0.03) & (0.03) &(0.03)
& (0.05) &(0.04) &(0.05) &(0.06) & (0.05) & (0.06) \\

Llama3.2-1B-ft vs. \hitgnn
& -0.07*** & -0.06** & -0.05** & 0.02 & -0.03 & -0.12***
& -0.03 & -0.03 & -0.01 & -0.05*& -0.02  & -0.05* \\
& (0.02) & (0.02) & (0.02) &(0.03) & (0.03) & (0.03)
& (0.03) & (0.02) &(0.03) &(0.03) & (0.03) &(0.03) \\

CLSTM vs \hitgnn& 
-0.04* &-0.02 &-0.02 &0.0 &-0.01 & -0.03& 
-0.04* &-0.02 &-0.02&-0.04&-0.02&0.00\\
&(0.02)&(0.02) &(0.02) & (0.02) &(0.02) &(0.03)
&(0.03)&(0.02)&(0.03)&(0.03)&(0.02)&(0.03)\\

\bottomrule
\end{tabular}
}
\caption{Pairwise model performance differences for top four models overall (first model minus second), based on bootstrap resampling test. Statistical significance: * $p<0.05$, ** $p<0.01$, *** $p<0.001$.}
\label{tab:statistical}
\vspace{-3mm}
\end{table*}
Table \ref{tab:phsample_entity} shows the frequency of annotated entities and relations. This annotation effort required $\approx16$ person-hours each for entity and relation annotation, resulting in $1{,}857$ entities and $279,560$ relation pairs (Appendix \ref{app:ph_annotation}). Table \ref{tab:coref} shows the performance evaluation of coreference resolution with entity linking using Metamap and Microsoft Recognizers-Text. While the performance on DATE is quite good, there is room for improvement across types. Except for TEST entities, the precision is high with lower recall, indicating the predicted clusters are more fragmented compared to the gold ones. The performance on selected entity types is higher than others, such as EVIDENTIAL, OCCURRENCE, which were not mapped in the \hitgnn pipeline as they cause further noise. For example, a matching string `reports' will be considered coreferential by Metamap, whether it refers to the same reporting incidence (e.g., the same reporter or what is being reported). A manual inspection of clusters from 150 notes concurs with these observations and reveals some sources of error. E.g., recurrence of a symptom doesn't indicate coreference, but is marked so. Recognizers-text is unable to normalize entries such as `pod \#3',  `yesterday', `this morning', with reference to the note date. Nevertheless, this linking still leads to improvement on the downstream tasks and provides a reliable foundation for further refinement.

\section{Statistical Significance}
\label{app:significance}

To assess whether the observed differences in performance metrics presented in Table \ref{tab:dp_results} are statistically significant, we perform non-parametric bootstrap resampling and compare each pair of the top models\textemdash CLSTM, \reveal, \hitgnn, and Llama3.2-1B-ft. Specifically, for each metric (e.g., macro-$F_1$, AUC, precision, recall), we repeatedly sample subsets of the test set with replacement and compute the performance difference between model pairs over 10,000 bootstrap replicates. This generates an empirical distribution of differences for each metric and model pair. From this distribution, we computed the mean difference, the sample standard deviation, and a 95\% confidence interval (CI) using the $2.5^{th}$ and $97.5^{th}$ percentiles. Statistical significance is determined based on the proportion of bootstrapped differences falling below or above zero, yielding a two-tailed p-value. Significance levels are reported using the convention: p<0.05 (*), p<0.01 (**), and p<0.001 (***). This approach makes no parametric assumptions and robustly quantifies uncertainty in model comparisons. The results are summarized in Table \ref{tab:statistical}.

\section{Ablations using T2D Recall}
\begin{figure*}
    \centering
\includegraphics[width=.9\linewidth]{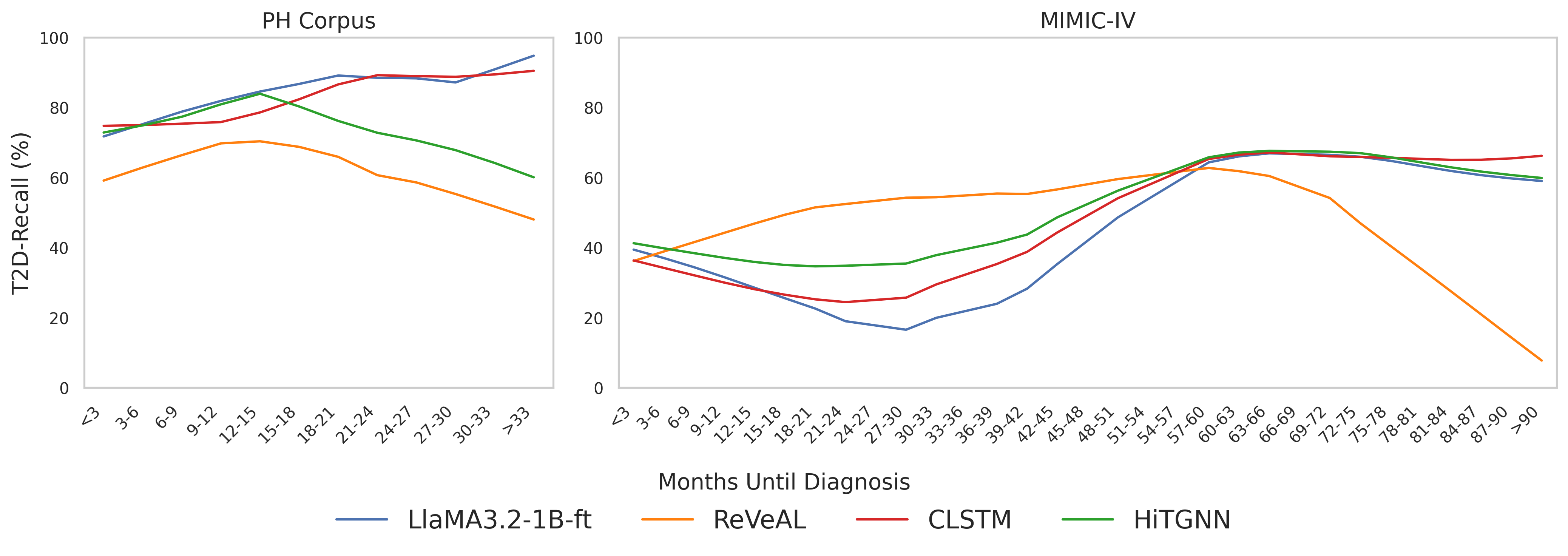}    
    \caption{T2D recall across consecutive 3-month prediction horizons.}
    \label{fig:dtd-ablation_recall}
    \vspace{-5mm}
\end{figure*}
\label{app:ablation_recall}
\paragraph{Prediction Horizons.} Figure \ref{fig:dtd-ablation_recall} presents the trends for T2D group recall across the top four models (Llama3.2-1B-ft, \reveal, CLSTM and \hitgnn with varying prediction windows. \hitgnn performs comparably in the immediate prediction window for the PH corpus, and has more stable performance across horizons for MIMIC-IV. \reveal has poor performance on the PH corpus but shows better performance in the medium horizon (9 months-4.5 years) for MIMIC-IV.

\begin{figure}
    \centering
    \begin{subfigure}[b]{0.5\linewidth}
        \centering
        \includegraphics[width=\linewidth, trim=20 12 10 5, clip]{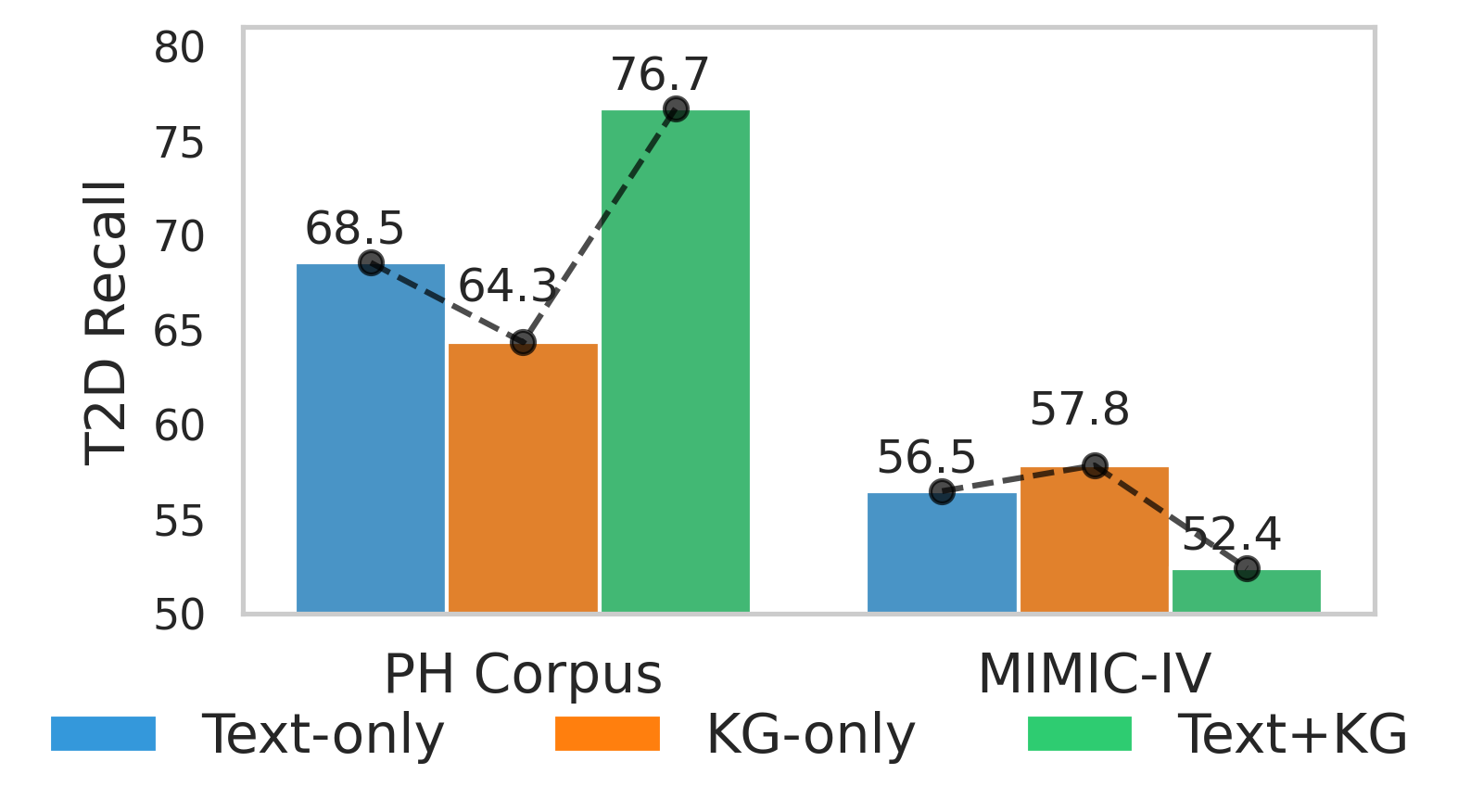}
        \caption{Embedding ablations.}
        \label{fig:node_embedding_ablations_t2d}
         \vspace{-3mm}
        \end{subfigure}
    \hfill
    \begin{subfigure}[b]{0.47\linewidth}
        \centering
        \includegraphics[width=\linewidth, trim=25 12 5 0, clip]{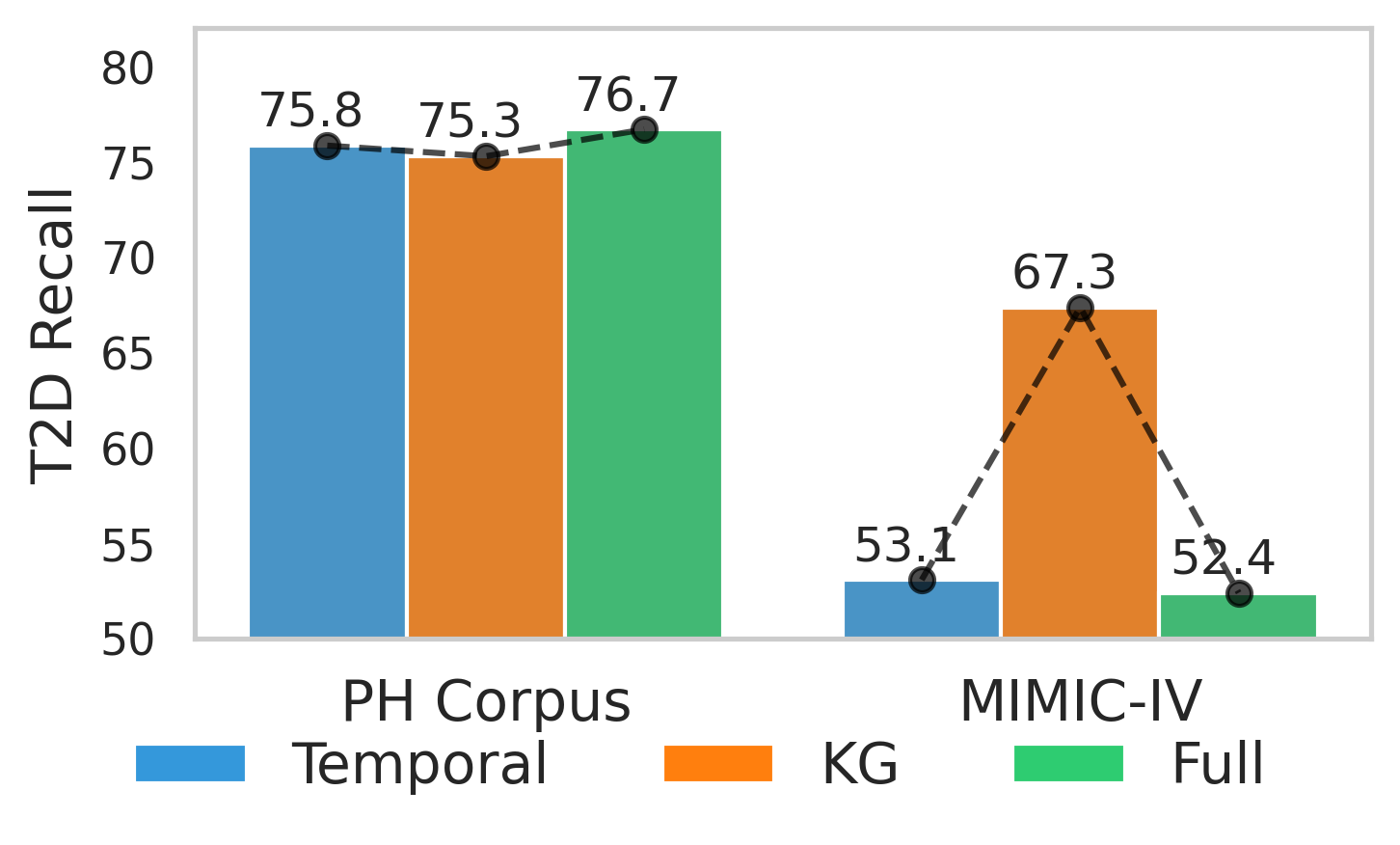}
        \caption{Subgraphs ablations.}
        \label{fig:subgraphsgnn_t2d}
         \vspace{-3mm}
    \end{subfigure}
    \caption{ \hitgnn ablations with T2D Recall.}
    \label{fig:hitgnn_ablations_t2d}
        \vspace{-2mm}
\end{figure}

\paragraph{Node Initialization and Subgraph Ablations. }Figures \ref{fig:node_embedding_ablations_t2d} and \ref{fig:subgraphsgnn_t2d} show the T2D recall trends across embedding and subgraph variations. In the PH corpus, text+KG node initialization consistently improves AUC and T2D recall, while KG-only edges yield the highest AUC; both temporal+KG improve T2D recall. In contrast, for MIMIC-IV, while both node embeddings and subgraphs lead to the highest AUC, KG-only node initialization and KG-only edges yield the highest T2D recall at the standard decision threshold (0.5). 

\section{LLM Fairness with and without Demographics}
\label{app:fairness}
\begin{table}[htbp]
\centering
\adjustbox{max width =1\linewidth}{
\begin{tabular}{llcccccc}
\toprule
&& \multicolumn{3}{c}{\textbf{PH Corpus}}& \multicolumn{3}{c}{\textbf{MIMIC-IV}}\\
\cmidrule(r){3-5}\cmidrule(r){6-8}
&&\textbf{G} & \textbf{Dem} & \textbf{no Dem}&\textbf{G} & \textbf{Dem} & \textbf{no Dem} \\
\midrule
\multirow{4}{*}{\rotatebox[origin=c]{90}{Ethnicity}}&\multirow{2}{*}{\rotatebox[origin=c]{90}{DPD}}
&H&   -0.05 & -0.05 & &&\\
&&N &0.05 & 0.05	&&&\\
\cmidrule{2-5}
&\multirow{2}{*}{\rotatebox[origin=c]{90}{EOD}}
&H& -0.06 & -0.07 & &\\
&&N &  0.06 & 0.07 	&&&\\
\midrule
\multirow{4}{*}{\rotatebox[origin=c]{90}{Gender}}&\multirow{2}{*}{\rotatebox[origin=c]{90}{DPD}}
&F &  0.05 & -0.03& F & -0.01 & -0.04 \\
&&M & -0.05 & 0.03 &M & 0.01 & 0.04 \\
\cmidrule{2-8}
&\multirow{2}{*}{\rotatebox[origin=c]{90}{EOD}}
&F& 0.05 & 0.02 & F & -0.04 & -0.07 \\
&&M & -0.05 & -0.02 & M &0.04 & 0.07  \\
\midrule
\multirow{8}{*}{\rotatebox[origin=c]{90}{Race}}&\multirow{4}{*}{\rotatebox[origin=c]{90}{DPD}}
&B& 0.09 & 0.01 & B &   0.04 & -0.12 \\
&& W &-0.07 & -0.02 & W &-0.06 & 0.10  \\
&&NI&  -0.07 & -0.01  & H &0.04 & -0.01\\
&&&&&A& 0.09 & 0.00 \\
\cmidrule{2-8}
&\multirow{4}{*}{\rotatebox[origin=c]{90}{EOD}}
&B& 0.13 & 0.03  & B &-0.03 & -0.23 \\
&&W& -0.08 & -0.03& W &  -0.08 & 0.12\\
&&NI&  -0.10 & -0.02 & H &0.09 & 0.09 \\
&&&&& A & 0.41 & 0.06 \\
\bottomrule
\end{tabular}
}
\caption{DPD/EOD for Llama3.1-8B with and without demographic conditioning.}
\label{tab:dem-nodem-Llama}
\end{table}
\section{Clinical Evaluation of \reveal Explanations}
\label{app:human-eval}
\paragraph{Case 1}: Expert-\reveal agree against ground truth.
\begingroup
\setstretch{0.95}
\setlength{\tabcolsep}{4pt}
\renewcommand{\arraystretch}{0.95}
\small
\begin{tabular}{@{}p{0.30\columnwidth}p{0.25\columnwidth}p{0.20\columnwidth}@{}}
\toprule
\textbf{Ground Truth} &\textbf{Expert}& \textbf{\reveal}\\
\midrule
Low & High & High \\
\midrule
\multicolumn{3}{p{\columnwidth}}{\textbf{Expert Rationale}:
Family history of diabetes and existing obesity indicate an elevated risk.
} \\
\midrule
\multicolumn{3}{p{\columnwidth}}{\textbf{\reveal Explanation}: The patient has a BMI of 33.4 kg/m\textsuperscript{2} which falls into the obese category (Class 2 obesity). Her BMI is consistently high across the two visit notes. Obesity is a significant risk factor for the development of Type 2 Diabetes, as it is associated with insulin resistance and other metabolic disturbances that contribute to the disease. The patient's family history also includes a first-degree relative (mother) with diabetes mellitus type 2, which further increases her risk.
} \\
\bottomrule
\end{tabular}
\endgroup
\paragraph{Case 2}: Expert-\reveal Agree with Ground Truth.
\begingroup
\setstretch{0.95}
\setlength{\tabcolsep}{4pt}
\renewcommand{\arraystretch}{0.95}
\small
\begin{tabular}{@{}p{0.30\columnwidth}p{0.25\columnwidth}p{0.20\columnwidth}@{}}
\toprule
\textbf{Ground Truth} & \textbf{Expert} & \textbf{\reveal} \\
\midrule

High& High & High  \\
\midrule
\multicolumn{3}{p{\columnwidth}}{\textbf{Expert Rationale}: Combination of African American, age, HTN, and high cholesterol raises my concern.
} \\
\midrule
\multicolumn{3}{p{\columnwidth}}{\textbf{\reveal Explanation}: The patient is a [AGE-Redacted]-year-old black female with a history of hypertension and hyperlipidemia, both of which are significant risk factors for Type 2 diabetes. Additionally, the patient is taking pravastatin, a medication commonly used to treat high cholesterol, which is often prescribed to individuals at risk for Type 2 diabetes. While the patient's current weight is not mentioned, her use of medications for hypertension and hyperlipidemia, in conjunction with her age and ethnicity, contributes to a higher risk for developing Type 2 diabetes. 
} \\
\bottomrule
\end{tabular}
\endgroup
\paragraph{Case 3a: Expert-\reveal Disagree}
\small
\noindent
\begin{tabular}{%
p{0.3\columnwidth}  
p{0.25\columnwidth}  
p{0.2\columnwidth}
}
\toprule
\textbf{Ground Truth} &\textbf{Expert}& \textbf{\reveal}\\
\midrule
Low & Low& High \\
\midrule
\multicolumn{3}{p{\columnwidth}}{\textbf{Expert Rationale}: She has multiple factors elevating her risk, HTN, and some other comorbidities, but she is still relatively young. I agree with AI's explanation that she is at increasing risk, but I would not say she is at high risk yet. Not sure about the concern about pseudoephedrine. Also ``Hypertriglyceridemia**: The patient's glucose level is elevated...potential insulin resistance'' is inaccurate.
} \\
\midrule
\multicolumn{3}{p{\columnwidth}}{\textbf{\reveal Explanation}: The patient's medical history and laboratory results from YEAR indicate several risk factors for Type 2 diabetes. 1. **Hypertriglyceridemia**: The patient's glucose level is elevated at 106 mg/dL (normal range: 70-99 mg/dL). While this is not extremely high, it is a subtle indication of potential insulin resistance. 2. **High CK levels**: The patient's creatine kinase (CK) level is elevated at 213 U/L (normal range: 24-195 U/L). Elevated CK levels can be a marker of metabolic syndrome, which is a cluster of conditions that increase the risk of developing Type 2 diabetes and cardiovascular disease. 3. **Hypertension**: The patient has a history of hypertension (HTN) and was treated with medication during her ED visit. Hypertension is a well-established risk factor for Type 2 diabetes. 4. **Family Medicine note**: The patient was instructed to stop taking pseudophedrine-containing products, which can raise blood pressure and potentially increase the risk of Type 2 diabetes. 5. **Age**: While age is a risk factor for Type 2 diabetes, the patient is 40 years old, which is relatively young for this diagnosis. However, combined with the other factors mentioned, it still increases her risk. 6. **Asthma and cough**: The patient has a history of asthma and was experiencing a cough at the time of her ED visit. These conditions may indicate an underlying inflammatory process that could be related to insulin resistance. In conclusion, while the patient's glucose level is not extremely high, the combination of elevated CK levels, hypertension, and other risk factors suggests that she is at a higher risk for developing Type 2 diabetes in the near future.
} \\
\bottomrule
\end{tabular}
\vspace{-5mm}
\paragraph{Case 3b: Expert-\reveal Disagree}
\small
\noindent
\begin{tabular}{%
p{0.3\columnwidth}  
p{0.25\columnwidth}  
p{0.05\columnwidth}  
}
\toprule
\textbf{Ground Truth} & \textbf{Expert}& \textbf{\reveal} \\
\midrule

High & High &Low \\
\midrule
\multicolumn{3}{p{\columnwidth}}{\textbf{Expert Rationale}: Two strong risk factors, severe obesity and Hispanic ethnicity, should be weighted more.
} \\
\midrule
\multicolumn{3}{p{\columnwidth}}{\textbf{\reveal Explanation}: The patient has a history of obesity, as documented in his notes from YEAR1 and YEAR2. However, despite his obesity, his laboratory results from YEAR2 show that his HbA1c (glycated hemoglobin) level is not elevated, indicating that his blood sugar control is within a normal range. Additionally, there is no mention of any other risk factors for Type 2 Diabetes, such as a family history of diabetes, hypertension, or a history of gestational diabetes. Furthermore, the patient's BMI, which was measured in YEAR2, is 55.7 kg/m\textsuperscript{2}, which is severely obese, but not significantly higher than his previous measurement. Given the absence of other risk factors and the patient's normal HbA1c level, the likelihood of him being diagnosed with Type 2 Diabetes in the near future is low. However, it is essential to continue monitoring his blood sugar levels and body weight to assess any potential changes in his risk profile. 
} \\
\bottomrule
\end{tabular}
					
\end{document}